\documentclass[runningheads]{llncs}

\usepackage{eccv}

\usepackage{eccvabbrv}

\usepackage{graphicx}
\usepackage{booktabs}

\usepackage[numbers,sort&compress]{natbib}
\definecolor{cvprblue}{rgb}{0.21,0.49,0.74}
\usepackage[pagebackref,breaklinks,colorlinks,allcolors=cvprblue]{hyperref}
\usepackage{tabularx}
\usepackage[skip=3pt]{caption} 
\usepackage{array}
\usepackage{colortbl}
\usepackage{cite}

\usepackage{tabularray}
\usepackage{tcolorbox}
\usepackage{multicol}
\usepackage{multirow}
\usepackage{array}     
\usepackage{calc}      
\usepackage{adjustbox}
\usepackage{pifont}
\usepackage{xcolor}
\usepackage{subcaption}
\usepackage{booktabs}
\usepackage{makecell}   
\usepackage{fontawesome}
\usepackage{wrapfig}
\usepackage{enumitem}
\usepackage{booktabs}    
\usepackage{xcolor}      
\usepackage{cancel}

\usepackage{amssymb}
\newtheorem{assumption}{Assumption}

\newcommand{\TARA}{{\tt TARA }}
\newcommand{\mycitet}[2]{#1~et~al.~\cite{#2}}

\usepackage[accsupp]{axessibility}  

\usepackage{hyperref}

\usepackage{orcidlink}

\usepackage{placeins}

\makeatletter

\newcommand{\printappendixtoc}{%
  \begingroup
  \noindent{\large\textbf{Table of Contents}}\par\medskip

  \def\l@section##1##2{%
    \begingroup
    \renewcommand{\numberline}[1]{\makebox[1.9em][l]{####1}}%
    \noindent
    \hbox to \linewidth{%
      \textbf{##1}\dotfill\textbf{##2}%
    }%
    \vspace{1mm}
    \par
    \endgroup
  }%

  \def\l@subsection##1##2{%
    \begingroup
    \renewcommand{\numberline}[1]{\makebox[2.8em][l]{####1}}%
    \noindent
    \hspace{1.5em}%
    \hbox to \dimexpr\linewidth-1.5em\relax{%
      ##1\dotfill##2%
    }%
    \vspace{1mm}
    \par
    \endgroup
  }%

  \@starttoc{atoc}%
  \endgroup
}

\newcommand{\asection}[1]{%
  \section{#1}%
  \addcontentsline{atoc}{section}{\protect\numberline{\thesection}#1}%
}

\newcommand{\asubsection}[1]{%
  \subsection{#1}%
  \addcontentsline{atoc}{subsection}{\protect\numberline{\thesubsection}#1}%
}

\makeatother

\newcommand{\mycolorbox}[2]{%
  \begingroup
  \setlength{\fboxsep}{1.3pt}%
  \colorbox{#1}{#2}%
  \endgroup
}

\begin{document}

\title{
Adapting MLLMs for Nuanced Video Retrieval
} 
\titlerunning{Nuanced Video Retrieval}

\author{Piyush Bagad\orcidlink{0000-0002-7533-5337} \and
Andrew Zisserman\orcidlink{0000-0002-8945-8573}}

\authorrunning{Bagad and Zisserman}

\institute{VGG, Department of Engineering Science\\
University of Oxford \\
{\footnotesize \url{https://www.robots.ox.ac.uk/~vgg/research/tara}}
}

\maketitle

\begin{abstract}
Our objective is to build an embedding model that captures the nuanced relationship between a search query and candidate videos. We cover three aspects of nuanced retrieval: (i) temporal, (ii) negation, and (iii) multimodal. For temporal nuance, we consider \textit{chiral actions} that need distinguishing between temporally opposite actions like ``opening a door'' \vs ``closing a door''. For negation, we consider queries with negators such as ``not'', ``none'' that allow a user to specify what they do not want. For multimodal nuance, we consider the task of \textit{composed retrieval} where the query comprises a video along with a text edit instruction. The goal is to develop a unified embedding model that handles such nuances effectively.
To that end, we repurpose a Multimodal Large Language Model (MLLM) trained to generate text into an embedding model. We fine-tune it with a contrastive loss on \textit{text alone} with carefully sampled hard negatives that instill the desired nuances in the learned embedding space. Despite the text-only training, our method achieves state of the art performance on \textit{all} benchmarks for nuanced video retrieval. We also analyze how this improvement is achieved, and  show that text-only training reduces the \textit{modality gap} between text and video embeddings leading to better organization of the embedding space.
\end{abstract}
\section{Introduction}
\label{sec:intro}


\begin{figure}[ht]
    \centering
    \includegraphics[width=\linewidth]{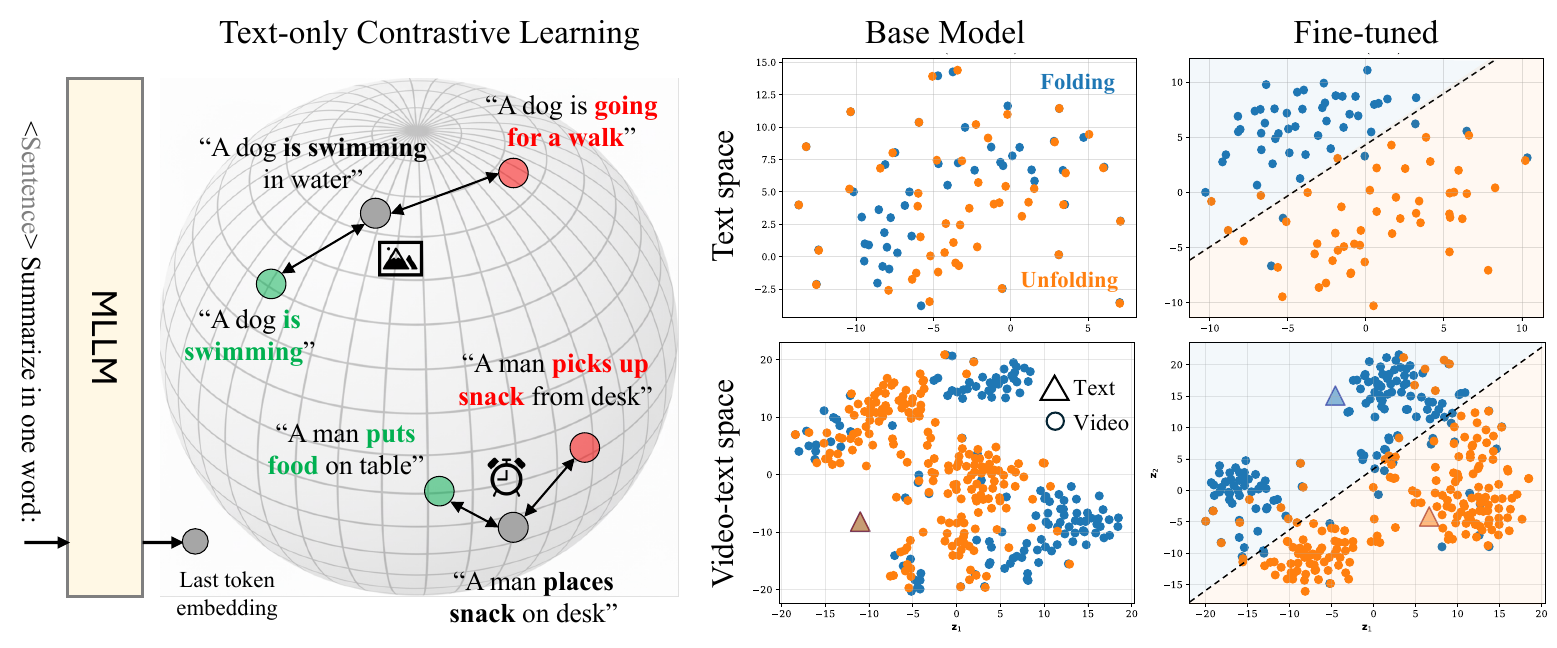}
    \caption{We repurpose an MLLM as a joint video-text embedding model. (Left) We fine-tune it on text samples with contrastive objective but with hard negatives chosen carefully to instill nuance in retrieval, \eg, temporal nuance (shown in the figure with a clock), to distinguish between \textit{chiral actions} that are temporally opposite in nature. (Right) We visualize the tSNE projections of embeddings before and after fine-tuning for a unseen action pair: \textit{folding} \vs \textit{unfolding} something. In the top row, our fine-tuned model better separates sentence embeddings specifying such pairs. The bottom row shows the joint video-text embedding space is adeptly organized for strong retrieval without training on videos. 
    In the Base Model, before fine-tuning, the text embeddings (shown in $\triangle$) for the two actions overlap in this visualization. After finetuning, the text embeddings are better separated and aligned with their video clusters.
    }
    \label{fig:teaser-with-tsne}
\end{figure}

The amount of video content on the Internet continues to grow rapidly with over 3M videos uploaded on YouTube every single day. Efficiently analyzing, organizing and searching through such scale is a necessity. Text provides a concise and efficient interaction layer between users and large-scale video content. Thus, developing reliable and performant video-text models for {\em retrieval} is crucial.
Furthermore, user search queries are often very specific and require {\em nuanced} video understanding. For example, consider the query ``Harry closes the door slowly, with nobody watching''. First, to perceive the door closing, the video model must take into account the frame-order, else it cannot distinguish between closing the door \vs opening the door. Likewise, it must also understand the pace of action, \ie, the door closes \textit{slowly}. Then, it should also understand negation. Finally, the user can augment the text query with an image/video of what Harry looks like. Each of these scenarios demands increasingly specific retrieved videos among a larger pool of related videos. We unify such challenges into the common theme of \textit{nuanced video retrieval}.

We consider three aspects of nuanced video retrieval. First, we consider time-sensitive retrieval where queries  involve some sort of temporal description and we want to retrieve videos that are \textit{temporally consistent} with the query among a set of videos that share similar spatial contexts but differ exactly in how they vary over time. Consider an example query, ``climbing up a ladder''.
We want to retrieve videos that show a person climbing up a ladder and not those of someone climbing down a ladder.
We measure this time-sensitivity on recently proposed benchmarks~\cite{bagad2025chirality,du2024reversed}. Second, we consider retrieval where the query is linguistically subtly modified by negation~\cite{alhamoud2025vision-negbench}.
The model must understand how such {\em negation} modifiers change semantics.
Third, we consider {\em multimodal} queries with the task of  composed video retrieval (CoVR)~\cite{ventura2024covr}. Here the model must understand how the text modifies (edits) the video example in forming the query for retrieval.

In this paper, we develop an encoder for the query and video such that they are mapped to a common embedding space but, unlike a dual encoder (\eg, CLIP), we use a unified encoder to produce both the query and the video embeddings by repurposing an MLLM. These embeddings are obtained by prompting an MLLM to summarize a given query/video and using the last token's final layer hidden representation as the embedding~\cite{jiang2024e5}, as illustrated in \cref{fig:teaser-with-tsne}.
Prior work has adapted MLLMs to generate embeddings suitable for retrieval, since the original 
MLLMs are typically trained for text generation on massive multimodal datasets, rather than for retrieval.  To adapt them for retrieval, a two-stage post-training approach is generally used~\cite{meng2025vlm2vec,li2026qwen3,liu2025lamra}: (i) first, the model is trained contrastively with positive and negative pairs from the text triplets of NLI~\cite{gao2021simcse} -- this adapts it for retrieval on text;
(ii) then, it is further trained on varying amounts of visual-text data to improve cross-modal retrieval.

The key insight of this paper is that by carefully choosing hard negatives to instill each of the three desired nuances: time, negation and multimodal, the MLLM can be fine-tuned using
\textit{only text triplets}.
For example in \cref{fig:teaser-with-tsne}, for temporal nuance (\faClockO), the anchor-positive share similar verb phrase while the hard-negative differs by using a temporally opposite \textit{chiral} verb phrase. Once trained, the model meaningfully encodes novel temporally opposite actions like ``folding / unfolding something'' as shown in the tSNE projections in \cref{fig:teaser-with-tsne}.
We show that by fine-tuning with this carefully curated training set of text triplets with hard negatives, we are able to achieve state of the art over multiple benchmarks for nuanced video retrieval -- this is quite surprising because the method surpasses others who have used far more training data, including a combination of text and visual.
We show that a partial reason for this success is that training with this curated set of text triplets 
closes the {\em modality gap}~\cite{liang2022mind,schrodi2025two} between the query and video embeddings.

In summary we make the following contributions: (i) We introduce a method for {\em Text Adapted Retrieval Alignment} ({\tt TARA}) that involves curating text-triplets with hard negatives for fine-tuning MLLMs to instill strong retrieval capabilities; (ii) We show that applying {\tt TARA} to any base MLLM {\em always} improves its performance on nuanced retrieval, often substantially, and indeed we achieve state of the art over multiple video retrieval benchmarks (including 
CiA~\cite{bagad2025chirality}, RTime~\cite{du2024reversed}, NegBench~\cite{alhamoud2025vision-negbench}, CoVR~\cite{ventura2024covr}); (iii) 
We show that text-only fine-tuning also improves performance on the video tasks of standard benchmarks like MMEB-V2~\cite{meng2025vlm2vec}; finally, (iv) We explain why text-only fine-tuning is so effective through the lens of the modality gap. 

It is worth noting that since the curated dataset is small, it only takes 
an hour to fine-tune a 7B model on 8 RTX A6000 GPUs.
All code, datasets and state of the art models are released on the \href{https://www.robots.ox.ac.uk/~vgg/research/tara}{project page}.

\section{Related Work}

\noindent\textbf{Video evaluation.}
Early work focused on action recognition with datasets like UCF~\cite{soomro2012ucf101}, HMDB~\cite{kuehne2011hmdb} and retrieval with MSRVTT~\cite{xu2016msr}, DiDeMo~\cite{anne2017localizing-didemo}. The dominance of MLLMs has prompted a suite of benchmarks for question-answering~\cite{xiao2021next,patraucean2023perception,li2024mvbench} and captioning~\cite{xu2024fine,wang2024tarsier,msvd}.
However, the community
has repeatedly discovered that most of these do not actually test for nuanced aspects, \eg time: a single frame or
an orderless set of frames can solve them~\cite{cores2024lost,huang2018makes,lei2023revealing,buch2022revisiting,zohar2025apollo,chen2024we,xue2025seeing}.
Most de-facto video retrieval datasets like MSRVTT~\cite{xu2016msr} also face this issue.
Meta-benchmarks like MMEB~\cite{meng2025vlm2vec,jiang2024vlm2vec} also inherit this issue as they are comprised of the same datasets.
Recent efforts~\cite{cores2024lost,xue2025seeing,saravanan2025velociti} aim to address this issue for video QA tasks.
\cite{chen2024beyond} proposed fine-grained evaluation with subtle single-word variations across nouns, verbs, adjectives, \etc.
In this work, we measure nuanced aspects in video retrieval through specialized benchmarks: \cite{du2024reversed,bagad2025chirality} for time, \cite{alhamoud2025vision-negbench} for negation, 
and \cite{ventura2024covr} for composed video retrieval.
\vspace{2mm}

\noindent\textbf{Fine-grained video retrieval.} 
Contrastively trained dual-encoder models like CLIP tend to focus on global, coarse-level vision-language matching~\cite{yuksekgonul2023when,chen2024beyond}. Early work on fine-grained retrieval explicitly modeled parts-of-speech into a joint multimodal space~\cite{wray2019fine}. \cite{Chen_2020_CVPR} proposed a Hierarchical Graph Reasoning model, which decomposes video-text matching into global-to-local levels by splitting text into embeddings for events, actions, entities. Recent work also adapts CLIP for fine-grained retrieval with densely annotated videos~\cite{ma2022xclip,singh2024figclip}. Other work adapt specialized models for specific aspects such as arrow of time~\cite{du2024reversed,xue2025seeing}, negation~\cite{wang2022learn}, adverb understanding~\cite{doughty2020action,doughty2022you}, verb understanding~\cite{momeni2023verbs}, \etc. In this work, we develop a unified recipe for fine-tuning an MLLM to instill some of these fine-grained qualities in video retrieval. 
\vspace{2mm}

\noindent\textbf{Adapting MLLMs for retrieval.}
We have witnessed a staggering rise in the abilities of open MLLMs on image~\cite{bai2025qwen2,deitke2025molmo,zhu2025internvl3} and video tasks~\cite{wang2024tarsier,bai2025qwen2,wang2025internvideo2}. A key benefit of open models is that we can analyze and use the hidden representations within the MLLM for retrieval. This has led to a new exciting area of adapting MLLMs as universal encoders~\cite{jiang2024e5,lin2024mmembed,meng2025vlm2vec,zhou2024vista,zhang2024gme,li2024improving-mcl}.
However, as also reflected in benchmarks for MLLMs like MMEB~\cite{meng2025vlm2vec}, the focus is still on images and static-biased video understanding.
Most contemporary work on adapting MLLMs for retrieval fine-tune on large-scale multimodal data using standard contrastive objective~\cite{li2026qwen3,meng2025vlm2vec,guo2025towards-gve,wei2024uniir}.
However, much like video-only datasets~\cite{lei2022revealing,buch2022revisiting}, these training datasets also suffer from focusing on static-biases over the nuances we care about.
Some early attempts at text-only fine-tuning~\cite{li2024topa,hong2026textme} map modality-specific encoders (\eg, CLIP for video) to a pre-trained LLM which makes it data-hungry (500K-4.2M training samples). In contrast, we directly adapt an MLLM post-trained for text generation into a retrieval model for temporal, negation and multimodal nuance with just 20K text triplets.
\vspace{2mm}

\noindent\textbf{Modality gap}, a systematic offset between image and text embeddings, was first identified in contrastive dual-encoder models like CLIP~\cite{liang2022mind}. Follow-up work \cite{zhang2023diagnosing,schrodi2025two} showed that this gap is constant across instances and classes, orthogonal to both modality subspaces. These insights led to methods showing that closing the gap (e.g., via mean centering \cite{MixedModalitySearch, yu2025unicorn, hong2026textme}) enables text-only training for tasks that typically need paired multimodal data \cite{C3,yu2026modality}.
\mycitet{Jiang}{jiang2024e5} 
argue that using the careful prompt while extracting embedding from an MLLM dissolves the modality gap which enables training on text alone.
They train on the NLI dataset~\cite{gao2021simcse} that has 275K text triplets.
Related work on videos also involves training on text but is complemented with training on videos either before~\cite{xu2024fine} or after~\cite{liu2025lamra} text training.
More recently, attention has turned to the modality gap in multimodal large language models (MLLMs). Methods based on preference/embedding alignment \cite{guiding-cross-modal,yin-etal-2025-sea} show that reducing the gap in MLLMs yields clear benefits for retrieval, suggesting it remains a key bottleneck even in unified architectures. We show that a simple text-only fine-tuning of an MLLM reduces the modality gap and better organizes the multimodal embedding spaces.
\section{Method: Text Adaptive Retrieval Aligment (TARA)}

Our goal is to adapt an MLLM $\mathcal{M}$ for video retrieval. Given a query $\mathbf{q}$ and video candidates $\{\mathbf{c}_{1}, \mathbf{c}_{2}, \ldots, \mathbf{c}_{N}\}$, we want to rank the candidates by relevance to the query using $\mathcal{M}$. Let $f_{\mathcal{M}}(.): \mathcal{X} \rightarrow \mathbb{R}^{d}$ denote a function to extract an embedding out of $\mathcal{M}$ where $\mathcal{X}$ is the space of all inputs and $d$ is the embedding dimension. Then, we can compute similarity $s$ between $\mathbf{q}$ and $\mathbf{c}$ as:
\begin{equation}
    s(\mathbf{q}, \mathbf{c}) = f_{\mathcal{M}}(\mathbf{q}). f_{\mathcal{M}}(\mathbf{c})^{T}
\end{equation}
Since $\mathcal{M}$ is only trained to generate text, the challenge is two-fold: (i) design $f_{\mathcal{M}}$, and (ii) fine-tune $\mathcal{M}$ such that $f_{\mathcal{M}}$ encodes desired semantic nuances.

\subsection{Designing $f_{\mathcal{M}}$: Extracting embeddings from MLLMs}
\label{subsec:review}

Following the work of
~\mycitet{Jiang}{jiang2024scaling-sentemb,jiang2024e5}
we use prompts to encourage the MLLM to condense the semantic meaning of a sentence or video into the hidden state of the next token, which is used as
the sentence or video embedding. This is termed an `Explicit One-word Limitation' (EOL) prompt. 
For example, an EOL prompt for a video is constructed as: ``\texttt{\small <video>: Summarize the video in one word: }''. Then, in generating the next token, the final layer hidden representation is used as the video embedding. This extends over the E5-V work of~\cite{jiang2024e5} where only images, text, and their combination were used. For a multimodal input (a video with a text edit instruction), we use a slightly different EOL prompt: ``\texttt{\small Source video: <video>; \ Edit instruction: <text>; Imagine this edit inst-\\ruction being applied to the source video. Summarize the resulting edited\\ video in one word:}''.
In our notation, this EOL-based embedding extraction process represents $f_{\mathcal{M}}$.

\subsection{Training $\mathcal{M}$: Text-only training enables cross-modal retrieval}

Text-only fine-tuning is straightforward: for a triplet $(\mathbf{t}_{i}, \mathbf{t}^{+}_{i}, \mathbf{t}^{-}_{i})$ where $\mathbf{t}^{+}_{i}$ is the embedding of a positive match sentence for the anchor $\mathbf{t}_{i}$, and $\mathbf{t}^{-}_{i}$ a hard-negative, the contrastive loss is given by:
\begin{equation}
\label{eq:loss}
    \mathcal{L}_{\text{con.}}(\mathcal{M}) = -\log \left( \frac{ e^{ \operatorname{s}(\mathbf{t}_{i}, \mathbf{t}^{+}_{i}) /\tau} }{ 
    \sum_{j}  e^{ \operatorname{s}(\mathbf{t}_{i}, \mathbf{t}^{+}_{j}) /\tau} + e^{ \operatorname{s}(\mathbf{t}_{i}, \mathbf{t}^{-}_{j}) /\tau}  
    } \right),
\end{equation}
where $\operatorname{s}(\mathbf{x}, \mathbf{y})$ denotes cosine similarity between $\mathbf{x}, \mathbf{y}$.
As will be shown in ~\cref{sec:analysis}, this text-only contrastive training reduces the modality gap between video and text embedding spaces. This leads to a meaningful re-organization of the embedding space which in-turn results in improved cross-modal retrieval. We refer to this text-only fine-tuning recipe as {\tt TARA}: Text Adapted Retrieval Alignment.

\subsection{Instilling nuance in retrieval with hard negatives}

\begin{figure}[ht]
    \centering
    \includegraphics[width=\linewidth]{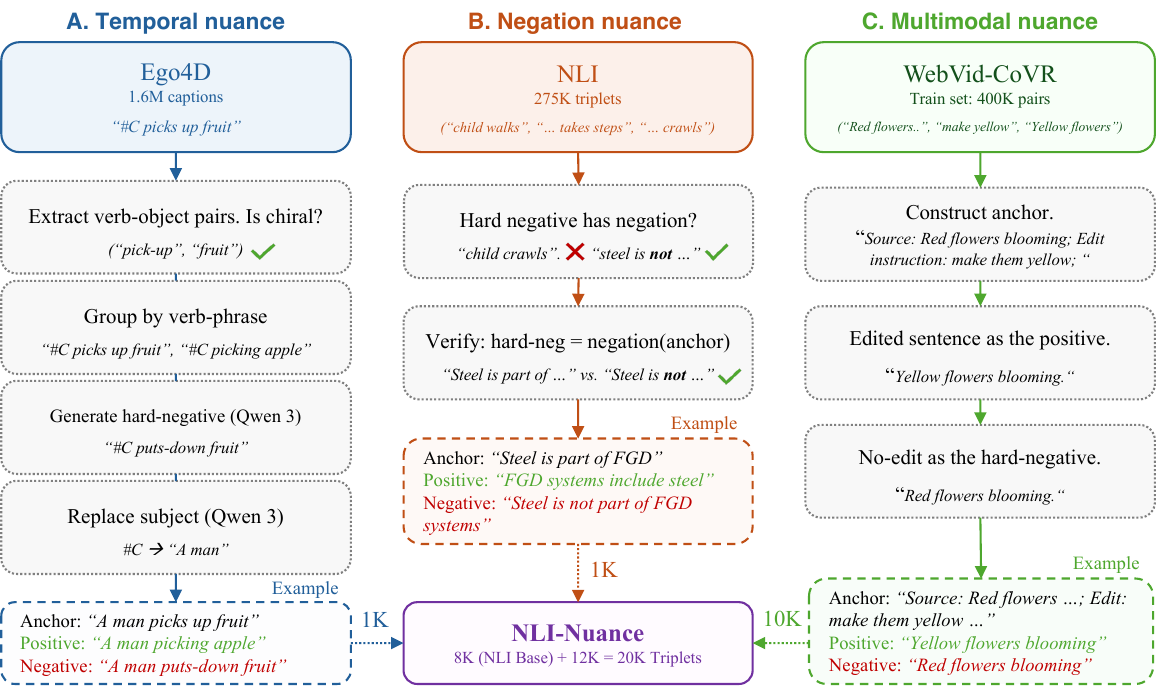}
    \vspace{-1mm}
    \caption{\textbf{Text dataset construction.} We augment samples in NLI with text triplets designed specifically to instill desired nuances in the embedding model. We call the resulting text dataset as NLI-Nuance comprising of 8K triplets from NLI and 12K triplets covering temporal, negation and multimodal nuances. More examples in \cref{subsec:ex-tripl}.}
    \label{fig:nli-nuance}
    \vspace{-5mm}
\end{figure}

Previous work~\cite{jiang2024e5,xu2024fine,liu2025lamra} has used text triplets from NLI~\cite{gao2021simcse} where ``entailment'' pairs are treated as positives and ``contradiction'' pairs as hard negatives. We argue that this helps with image understanding but is insufficient to instill nuance in retrieval.
Consider the triplet (``child \textcolor{blue}{walking on concrete ledge}'', ``\ldots \textcolor{blue}{takes steps along the concrete}'', ``\ldots \textcolor{red}{crawls along sandy ledge}''). The positive and anchor describe a child walking on concrete while the negative differs in \textit{crawling on sandy edge}.
If these were captions of videos, a single frame would suffice to recognize the positive to be closer to the anchor, with no need for nuanced temporal understanding.
Thus, we augment NLI with text triplets specifically designed to instill nuanced understanding in retrieval for (i) temporal, (ii) negation and (iii) multimodal. We sketch the steps to generate hard-negatives in \cref{fig:nli-nuance}.
\vspace{-2mm}
\begin{enumerate}[label=\Alph*.]
    \item \small \textbf{Temporal nuance.} We extract verb-object pairs in Ego4D~\cite{grauman2022ego4d} and pick only those with a \textit{chiral} verb~\cite{bagad2025chirality} based on a pre-compiled list of chiral verbs generated by Claude. Positives are chosen s.t. they share the chiral verb with the anchor while hard-negatives are generated by prompting Qwen 3 
    as shown with the example in \cref{fig:nli-nuance}.
    In \cref{subsec:temp-anto}, we detail verification of the hard-negatives via LLM-as-a-Judge followed by a manual review.
    \item \textbf{Negation nuance.} We filter those triplets from NLI where the hard-negative has a negation operator (\eg, ``not''). Example is shown in \cref{fig:nli-nuance}.
    \item \textbf{Multimodal nuance.} Building on the idea of converting a video task to a text-only task, we translate Composed Video Retrieval to a Composed Text Retrieval. Starting with train set of WebVid-CoVR~\cite{ventura2024covr}, we replace the source video and to-be-retrieved video with their corresponding captions as shown in \cref{fig:nli-nuance}.
\end{enumerate}
\vspace{-2mm}

\noindent In order to retain the static understanding of the model, we also retain text triplets from the original NLI dataset. In summary, we build a text dataset called \textbf{NLI-Nuance} consisting of a total of $N{=}20,000$ triplets. We fine-tune a given MLLM on this dataset. Note that this dataset consists of text samples sourced from Ego4D train set, WebVid-CoVR train set and the original NLI dataset. {\em It does not use any text captions from the evaluation datasets.}
\vspace{-2mm}
\section{Experimental Results}
\vspace{-1mm}

Our experiments are designed to measure if \TARA generalizes to nuanced video retrieval.
In the following Sections, we introduce evaluation datasets for each aspect of nuanced video retrieval. As each dataset may have a different evaluation metric, we introduce these as we describe the dataset.
In \cref{subsec:time}, we evaluate temporal nuance on time-sensitive benchmarks for video retrieval. In \cref{subsec:linguistic}, we probe two kinds of linguistic nuance: negation and adverbs in text queries. In \cref{subsec:multimodal}, we test multimodal nuance through composed video retrieval.
In each case, we show that \TARA improves performance, sometimes significantly, and this applies across multiple open-source MLLMs.
In \cref{subsec:standard}, we show that \TARA does {\em not} harm retrieval on standard evaluation datasets.
For each evaluation dataset, we provide the number of query and number of candidates in \cref{tab:data-stats}. 
Ablations on training data composition and size are provided in \cref{asubsec:data}.
\vspace{2mm}

\noindent\textbf{Implementation details.}
While we experiment with 5 MLLMs (Qwen2VL~\cite{wang2024qwen2}, Qwen3VL~\cite{li2026qwen3}, InternVL3~\cite{zhu2025internvl3}, Tarsier 1~\cite{wang2024tarsier} and 2~\cite{yuan2025tarsier2}), we obtain best results with fine-tuning Tarsier 2 which we treat as our default model.  
We only fine-tune the LLM weights and freeze the vision and projection networks. We train for 2 epochs with batch size of 768 and base learning rate $2e{-}5$. On 8 Nvidia RTX A6000 GPUs, it takes less than an hour to train \TARA. During inference, we use $F{=}16$ uniformly spaced frames in the video.
\vspace{-5mm}
\begin{table}[h!]
\centering
\footnotesize
\caption{\textbf{Evaluation Dataset statistics.} 
}
\vspace{-4mm}
\resizebox{0.9\columnwidth}{!}{%
\begin{tabular}{lccc cc c}
\toprule
\textbf{} 
& \multicolumn{3}{c}{\textbf{CiA} (All)} 
& \multicolumn{2}{c}{\textbf{NegBench}} 
& \textbf{WebVid-CoVR} \\ 
\cmidrule(lr){2-4} \cmidrule(lr){5-6}
& SSv2 & EPIC & Charades 
& MS-COCO & MSRVTT 
& \\ 
\midrule
Number of Queries &  32 &  128  & 56 & 5915 & 995 & 2556 \\
Average number of Candidates &  1430 &  3108 & 5498 & 5915 & 1000 & 2556 \\
\bottomrule
\end{tabular}%
}
\label{tab:data-stats}
\vspace{-11mm}
\end{table}

\subsection{Temporal Nuance}
\label{subsec:time}
\vspace{-1mm}

\noindent\textbf{CiA evaluation dataset.} First, we use the Chirality in Action (CiA) dataset proposed by
~\mycitet{Bagad}{bagad2023testoftime}.
It is made up of temporal actions from three datasets: SSv2, EPIC and Charades. We use the corresponding text captions as queries and measure mAP over the retrieved videos. CiA has three settings: (i) \mycolorbox{yellow!15}{\textit{chiral}}: the video gallery $\mathcal{G}$ consists of only the correct action and its temporal opposite (\eg, for query `opening door', $\mathcal{G}$ has videos of opening and closing doors.), (ii) \mycolorbox{Lavender!20}{\textit{static}}: $\mathcal{G}$ consists of the correct action and all other temporally irrelevant actions (\eg,  for query `opening door', $\mathcal{G}$ has distractors `folding paper', `wrapping gift', \etc), (iii) \mycolorbox{Sepia!10}{\textit{all}}: $\mathcal{G}$ consists of the correct action and all other videos, \ie, it has the temporal opposite action as well as temporally irrelevant actions.
Results are presented in \cref{tab:cia-combined-t2v}. \TARA comprehensively achieves state of the art performance on CiA while being fine-tuned on text alone.
Note, TARA always improves over the base model for chiral action retrieval. The degree of improvement depends on the base model: Q3VL-Emb.~\cite{li2026qwen3} is already trained contrastively on multimodal data, so TARA shows smaller benefits. But with other base models trained autoregressively, TARA shows much larger benefits.

\vspace{-9mm}
\begin{table*}[h]
\centering
\caption{\small \textbf{Results on CiA-Retrieval.} We report mAP for T2V retrieval on CiA~\cite{bagad2025chirality}. 
\mycolorbox{yellow!15}{\textit{Chiral}}
denotes retrieving from a video gallery $\mathcal{G}$ of only temporally antonymous (chiral) samples
, \mycolorbox{Lavender!20}{\textit{Static}}: $\mathcal{G}$ has non-chiral actions and 
\mycolorbox{Sepia!10}{\textit{All}}: $\mathcal{G}$ includes chiral/non-chiral.
\mycolorbox{green!8}{\tt TARA} always improves over base model; achieves state-of-the-art performance with Tarsier-2.
}
\vspace{-3mm}
\resizebox{\linewidth}{!}{%
\small
\begin{tabular}{l c l ccc ccc ccc}
\toprule
\small \textbf{Method}
&
\multicolumn{2}{c}{\textbf{Fine-tuning data}} 
& \multicolumn{3}{c}{\textbf{SSv2}}
& \multicolumn{3}{c}{\textbf{EPIC}}
& \multicolumn{3}{c}{\textbf{Charades}} \\
\cmidrule(lr){2-3} \cmidrule(lr){4-6} \cmidrule(lr){7-9} \cmidrule(lr){10-12}
& Size (K) & Modalities & {\small \colorbox{yellow!15}{\textit{Chiral}}} & {\small \colorbox{Lavender!20}{\textit{Static}}} & {\small \colorbox{Sepia!10}{\textit{All}}} & {\small \colorbox{yellow!15}{\textit{Chiral}}} & {\small \colorbox{Lavender!20}{\textit{Static}}} & {\small \colorbox{Sepia!10}{\textit{All}}} & {\small \colorbox{yellow!15}{\textit{Chiral}}} & {\small \colorbox{Lavender!20}{\textit{Static}}} & {\small \colorbox{Sepia!10}{\textit{All}}} \\
\midrule
Chance & - & - & 50.0 & 6.3 & 3.1 & 50.0 & 1.5 & 0.8 & 50.0 & 3.6 & 1.8 \\
\arrayrulecolor{gray!70}\midrule\arrayrulecolor{black}
\multicolumn{12}{c}{\cellcolor[HTML]{e8f4fc}\textbf{Dual encoder models}}\\
CLIP (avg.)~\cite{clip-radford2021learning} & - & - & 52.0 & 18.3 & 12.7 & 51.0 & 12.0 & 7.0 & 48.4 & 13.2 & 6.5 \\
DINO.txt~\cite{dino.txt} & - & - & 52.1 & 19.9  & 13.1 & 50.6 & 14.6 & 6.0 & 50.7 & 19.0 & 10.1 \\
Perception Enc.~\cite{bolya2025perception} & - & - & 50.1 & 32.7 & 17.2 & 48.5 & 4.2 & 1.8 & 51.3 & 12.8 & 7.2 \\
InternVideo 2~\cite{wang2024internvideo2} & - & - & 52.5 & 35.7 & 20.6 & 48.3 & 22.1 & 8.8 & 50.7 & 11.9 & 11.9 \\
\arrayrulecolor{gray!70}\midrule\arrayrulecolor{black}
\multicolumn{12}{c}{\cellcolor[HTML]{e8f4fc}\textbf{MLLMs}}\\
VLM2Vec-V2~\cite{meng2025vlm2vec} & 1700 & {\footnotesize \faFilm \ \faPhoto \ \faPencil \ \faFilePdfO} & 58.8 & 27.8 & 15.9 & 49.4 & 25.4 & 12.9 & 53.5 & 18.8 & 10.5 \\
LAMRA~\cite{liu2025lamra} & 1400 & {\footnotesize \faPhoto \ \faPencil} & 55.3 & 15.2 & 7.8 & 53.7 & 11.3 & 9.0 & 52.1 & 21.2 & 11.3 \\
GVE-7B~\cite{guo2025towards-gve} & 13000 & {\footnotesize \faFilm \ \faPhoto \ \faPencil} & 53.4 & 17.1 & 4.0 & 54.7 & 22.4 & 7.3 & 54.2 & 21.0 & 10.2 \\
E5-V~\cite{jiang2024e5} & 300 & {\footnotesize \faPencil} & 52.6 & 18.9 & 14.7 & 57.1 & 16.4 & 6.5 & 48.9 & 19.7 & 7.1 \\
ArrowRL~\cite{xue2025seeing} & 25 & {\footnotesize \faFilm \ \faPencil} & 67.5 & 33.8 & 22.5 & 55.7 & 12.4 & 9.6 & 57.1 & 18.6 & 12.2 \\
CaRe~\cite{xu2024fine} & 275 & {\footnotesize \faFilm \ \faPencil} & 66.4 & 46.2 & 23.7 & 62.3 &  25.0 & 16.9 & 56.1 & 25.2 & 12.9 \\
Qwen2VL-7B~\cite{wang2024qwen2} & - & - & 60.2 & 28.0 & 17.3 & 53.7 & 11.2 & 9.6 & 55.9 & 16.7 & 8.1 \\
\rowcolor{green!8} Qwen2VL + \TARA & 20 & {\footnotesize \faPencil} & 70.1 & 39.6 & 27.4 & 65.0 & 27.9 & 20.8 & 65.5 & 31.8 & 22.7 \\
Qwen2.5VL-7B~\cite{bai2025qwen2} & - & - & 67.6 & 31.5 & 20.6 & 55.4 & 12.4 & 9.5 & 55.8 & 17.0 & 11.1 \\
Qwen3VL-7B~\cite{bai2025qwen3} & - & - & 56.8 & 16.4 & 11.0 & 57.8 & 2.9 & 1.8 & 52.1 & 9.1 & 6.6 \\
\rowcolor{green!8} Qwen3VL + \TARA & 20 & {\footnotesize \faPencil} & 63.6 & 23.5 & 13.6 & 64.0 & 25.1 & 20.4 & 59.4 & 17.5 & 12.4 \\
Qwen3VL-Emb.~\cite{li2026qwen3} & NA & {\footnotesize \faFilm \ \faPhoto \ \faPencil \ \faFilePdfO} & 72.0 & 43.4 & 31.8 & 62.1 & 28.6 & 20.6 & 65.3 & 37.3 & 26.1 \\
\rowcolor{green!8} Qwen3VL-Emb + \TARA \ \ & 20 & {\footnotesize \faPencil} & 73.0 & 51.5 & 37.8 & 64.0 & 25.1 & 20.1 & 68.3 & 36.4 & 28.7 \\
InternVL3-8B~\cite{zhu2025internvl3} & - & - & 56.8 & 12.5 & 6.5 & 57.5 & 6.2  & 4.3 & 54.0 & 16.1 & 9.4 \\
\rowcolor{green!8} InternVL3 + \TARA & 20 & {\footnotesize \faPencil} & 73.1 & 36.0 & 27.6 & 68.9 & 33.9 & 24.3 & 64.1 & 27.5 & 19.8 \\
Tarsier~\cite{wang2024tarsier} & - & - & 76.4 & 51.8 & 35.0 & 62.7 & 22.1 & 17.4 & 64.7 & 25.9 & 17.7 \\
\rowcolor{green!8} Tarsier + \TARA & 20 & {\footnotesize \faPencil} & 84.9 & 61.4 & 49.5 & 80.5 & 37.3 & 31.7 & 71.7 & 38.7 & 27.8  \\
Tarsier2-7B~\cite{yuan2025tarsier2} & - & - & 77.7 & 26.9 & 24.0 & 67.4 & 22.0 & 15.3 & 60.5 & 13.4 & 9.2 \\
\rowcolor{green!8} Tarsier 2 + \TARA & 20 & {\footnotesize \faPencil} & \textbf{88.9} & \textbf{66.7} & \textbf{58.6} & \textbf{81.1} & \textbf{45.6} & \textbf{38.9} & \textbf{71.4} & \textbf{38.6} & \textbf{29.0} \\
\bottomrule
\end{tabular}%
}
\label{tab:cia-combined-t2v}
\vspace{-9mm}
\end{table*}

\begin{wrapfigure}[12]{r}{0.39\textwidth}
\vspace{-10mm}          
\centering
\small
\setlength{\tabcolsep}{1.6pt}
\renewcommand{\arraystretch}{0.92}
\captionof{table}{R@1 on RTime.}
\label{tab:rtime}
\begin{tabular}{@{}lcc@{}}
\toprule
\textbf{Method} & \textbf{T2V} & \textbf{V2T} \\
\midrule
\multicolumn{3}{c}{\cellcolor[HTML]{e8f4fc}\textbf{Zero-shot}} \\
Singularity & 48.7 & 49.9 \\
Internvideo2-1B & 50.0 & 51.0 \\
Qwen2.5VL & 53.4 & 66.6 \\
Tarsier 2 & 58.8 & 59.5 \\
\rowcolor{green!8}
Tarsier 2 + \TARA & \textbf{67.2} & \textbf{77.9} \\
\midrule
\multicolumn{3}{c}{\cellcolor[HTML]{e8f4fc}\textbf{Fine-tuned on RTime}} \\
CLIP4Clip & 49.8 & 49.8 \\
UMT & 51.2 & 51.3 \\
UMT-Neg & 54.5 & 54.2 \\
ArrowR-Qwen2 & 57.1 & 68.8 \\
ArrowRL-Qwen2.5 & 55.6 & 69.6 \\
\bottomrule
\end{tabular}
\vspace{-9.5pt}
\end{wrapfigure}
\noindent\textbf{RTime.} We also evaluate on the Reversed in Time (RTime) benchmark by 
\mycitet{Du}{du2024reversed}
that uses the arrow of time to generate distractor videos: a video with caption `a woman opening a laptop' is time-reversed with a new caption `a woman closing a laptop'. 
For V2T, the task is to choose between the correct and reversed caption.
Likewise, for T2V, the task is to choose between the actual and reversed video. 
Results (R@1) are shown in \cref{tab:rtime}. 
\TARA outperforms all competing methods, including those from~\cite{du2024reversed} that are fine-tuned on the RTime training set.

\subsection{Negation Nuance}
\label{subsec:linguistic}

We use the NegBench~\cite{alhamoud2025vision-negbench} benchmark to evaluate text-to-image (\colorbox{yellow!15}{COCO}) and text-to-video (\colorbox{Orchid!10}{MSRVTT}) retrieval where text queries involve negation. We evaluate on standard queries (\eg, `a picture of a dog') and on negated queries (\eg, `a picture of a dog but \textit{not on grass}'). We compare methods from 
\mycitet{Alhamoud}{alhamoud2025vision-negbench}
fine-tuned on variations of Conceptual Captions (CC) and report R@5. As shown in \cref{tab:negbench}, \TARA outperforms all competing methods on both datasets.
Relatedly, we also show {\tt TARA}'s \textit{adverb} understanding in \cref{asubsec:adverbs}, \eg, given a video of a person walking, distinguishing if the walking is ``slow''/``quick''.
\vspace{-5mm}

\vspace{-3mm}
\begin{table}[h]
\centering
\caption{\textbf{NegBench Evaluation} checks understanding of negation in queries.
\TARA (zero-shot) beats strong baselines in~\cite{alhamoud2025vision-negbench}.}
\label{tab:negbench}
\footnotesize
\vspace{-3mm}
\begin{tabular}{@{}llcccc@{}}
\toprule
\textbf{Method} & \textbf{Fine-tuning data}
  & \multicolumn{2}{c}{\colorbox{yellow!15}{\textbf{COCO}}} 
  & \multicolumn{2}{c}{\colorbox{Orchid!15}{\textbf{MSR-VTT}}} \\
\cmidrule(lr){3-4}\cmidrule(lr){5-6}
 & & \small R@5 &  \small  R-Neg@5  & \small  R@5  & \small  R-Neg@5 \\
\midrule

\multirow{4}{*}{CLIP}
  & None & 54.8 & 48.0 & 50.6 & 45.8 \\
  & CC (\faPhoto, \faPencil) & 58.8 & 54.5 & 53.7 & 49.9 \\
  & CC-NegCap (\faPhoto, \faPencil) & 58.5 & 57.8 & 54.1 & 53.5 \\
  & CC-NegFull (\faPhoto, \faPencil) & 54.2 & 51.9 & 46.9 & 43.9 \\
\midrule

\multirow{4}{*}{NegCLIP~\cite{alhamoud2025vision-negbench}}
  & None & 68.7 & 64.4 & 53.7 & 51.0 \\
  & CC (\faPhoto, \faPencil) & 70.2 & 66.0 & 56.4 & 52.6 \\
  & CC-NegCap (\faPhoto, \faPencil) & 68.6 & 67.5 & 56.5 & 54.6 \\
  & CC-NegFull (\faPhoto, \faPencil) & 69.0 & 67.0 & 54.0 & 51.5 \\
\midrule

Tarsier 2 & None & 33.3 & 21.5 & 25.6 & 18.9 \\
\rowcolor{green!8}
Tarsier 2 + \TARA & NLI-Nuance (\faPencil) & \textbf{76.7} & \textbf{73.6} & \textbf{65.1} & \textbf{65.0} \\
\bottomrule
\end{tabular}

\end{table}

\vspace{-8.5mm}
\subsection{Multimodal Nuance}
\label{subsec:multimodal}

\begin{wraptable}[17]{r}{0.45\textwidth}
\centering
\vspace{-14mm}
\caption{\small \textbf{Composed video retrieval on WebVid-CoVR.} \TARA beats all competing methods while being fine-tuned only with text data.}
\resizebox{ \linewidth}{!}{%
\begin{tabular}{l r r r}
\toprule
Method & R@1 & R@5 & R@10 \\
\midrule
Chance & 0.1 & 0.2 & 0.4 \\
\midrule
\multicolumn{4}{c}{\cellcolor[HTML]{e8f4fc}\normalsize \textbf{Zero-shot}}\\
BLIP (T) & 19.7 & 37.1 & 45.9 \\
BLIP (V) & 34.9 & 59.2 & 68.0 \\
CLIP (V+T) & 44.4 & 69.1 & 77.6 \\
BLIP (V+T) & 45.5 & 70.5 & 79.5 \\
\rowcolor{green!8} Tarsier 2+\TARA & \textbf{66.3} & \textbf{86.7} & \textbf{91.5} \\
\midrule
\multicolumn{4}{c}{\cellcolor[HTML]{e8f4fc}\normalsize \textbf{Fine-tuned on CoVR}}\\
BLIP (T) & 23.7 & 45.9 & 55.1 \\
BLIP (V) & 38.9 & 65.0 & 74.0 \\
CLIP (V+T) & 50.6 & 77.1 & 85.1 \\
BLIP (V+T) & 50.6 & 74.8 & 83.4 \\
BLIP (V+T) & 51.8 & 78.3 & 85.8 \\
\rowcolor{orange!8} \mycitet{Ventura}{ventura2024covr} & {53.1} & {79.9} & {86.9} \\
\rowcolor{orange!8} \mycitet{Ventura}{ventura2024covr2} & {59.8} & {83.8} & {91.3} \\
\bottomrule
\end{tabular}
}
\label{tab:webvid-covr-results}
\vspace{-5mm}
\end{wraptable}
Since \TARA is based on extracting embeddings out of an MLLM, it inherits the flexibility of an MLLM to take as input the composition of video-text together. We leverage this and evaluate on the task of Composed Video Retrieval introduced by \mycitet{Ventura}{ventura2024covr}. The queries are composed of a video and a text edit instruction. Similar to the EOL prompt for each modality, we modify the EOL prompt slightly for joint video-text inputs. We provide the full prompts used in the Appendix.
We evaluate on the WebVid-CoVR~\cite{ventura2024covr} test set consisting of 2,556 query-video samples. \TARA is evaluated in a zero-shot manner.
As shown in \cref{tab:webvid-covr-results}, \TARA outperforms even methods fine-tuned on WebVid-CoVR.

\subsection{Standard Benchmarks}
\label{subsec:standard}

\vspace{-4mm}
Does text-only fine-tuning on our NLI-Nuance dataset deteriorate performance on standard benchmarks for video recognition/retrieval? To test this, we evaluate \TARA on video tasks from the MMEB-V2 benchmark~\cite{meng2025vlm2vec} across 10 standard datasets for classification and retrieval.
The results are tabulated in \cref{tab:mmebv2-subset}. We compare with the top multimodal embedding models~\cite{zhang2024gme,meng2025vlm2vec} including recently proposed thinking-augmented retrieval~\cite{cui2025think}.
First, we note that \TARA comprehensively improves upon its base model (Tarsier 2). Second, despite being fine-tuned only on text samples, it is competitive with other methods trained on massive amounts of multimodal data. In terms of average classification/retrieval performance, it is second only to Qwen3-VL-Embedding~\cite{li2026qwen3}.
Furthermore, we also explore a a simple ensemble (Concatenation of Embeddings) of TARA Qwen3-VL-Embeddings and show that it outperforms the latter,
demonstrating the complementary abilities that TARA possesses.

\begin{table}[h!]
\vspace{-5mm}
\centering
\caption{\textbf{Results on subset of tasks in MMEB-V2.} Despite fine-tuning only on text samples, \TARA comprehensively improves upon the base model and is only second to Qwen3VL-Embedding trained on massive amount of multimodal data.
TARA $\oplus$ Q3VLE denotes a concatenation
of TARA and Qwen3VL-Embedding. It outperforms
Qwen3VL on average highlighting the complementary abilities that TARA possesses.
}
\vspace{-3mm}
\resizebox{\columnwidth}{!}{%
\begin{tabular}{lllllllllllll}
\toprule
\textbf{Method} & \multicolumn{6}{c}{\textbf{Video classification}} & \multicolumn{6}{c}{\textbf{Video retrieval}} \\ 
\cmidrule(lr){2-7} \cmidrule(lr){8-13}
 & UCF & HMdb & K700 & BF & SSv2 & Avg. & MSR & MSVD & DDMo & YC2 & VTX & Avg. \\
 \midrule
Chance & 1.0  & 2.0 & 0.1 & 2.2 & 0.6 & - & 0.1 & 0.1 & 0.1 & 0.1 & 0.1 & -\\ 
ColPali v1.3 & 49.4 & 24.8 & 23.4 & 10.9 & 25.1 & 26.7 & 17.6 & 45.4 & 22.8 & 5.3 & 16.7 & 21.6 \\
GME-2B & 52.4 & 43.4 & 35.2 & 13.6 & 29.9 & 34.9 & 27.3 & 47.6 & 22.0 & 7.9 & 23.0 & 25.6 \\
GME-7B & 54.7 & 47.9 & 39.7 & 14.3 & 30.6 & 37.4 & 31.8 & 49.7 & 26.4 & 9.1 & 24.9 & 28.4 \\
LamRA-Qwen2 & 60.4 & 40.5 & 42.3 & 16.9 & 36.3 & 39.3 & 22.1 & 46.1 & 24.8 & 9.2 & 19.1 & 24.3 \\
LamRA-Qwen2.5~\cite{liu2025lamra} & 53.0 & 33.8 & 32.1 & 20.1 & 25.3 & 32.9 & 25.0 & 41.9 & 22.8 & 7.5 & 18.7 & 23.2 \\
VLM2Vec-2B & 57.5 & 33.8 & 31.4 & 13.4 & 30.9 & 33.4 & 25.2 & 38.2 & 19.4 & 4.1 & 16.2 & 20.6 \\
VLM2Vec-7B & 61.8 & 42.2 & 35.5 & 23.8 & 32.1 & 39.1 & 34.5 & 46.7 & 29.3 & 9.0 & 25.5 & 29.0 \\
VLM2Vec-V2.0~\cite{meng2025vlm2vec} & 60.0 & 40.9 & 38.0 & 14.8 & 42.8 & 39.3 & 28.3 & 48.1 & 30.4 & 10.6 & 26.5 & 28.8 \\
TTE-7B~\cite{cui2025think} 
& 78.6 & 63.9 & 55.6 & 34.2 & 55.3 & 57.5 
& 39.5 & 59.4 & 36.3 & \underline{20.3} & 32.6 & 37.6 \\
Tarsier 2 
& 37.9 & 17.4 & 29.6 & 36.1 & 15.9 & 27.4 
& 9.5 & 39.8 & 12.2 & 3.9 & 16.6 & 16.4 \\
\rowcolor{green!8} 
Tarsier 2 + \TARA (Ours) 
& 80.3 & 69.0 & 59.4 & 45.6 & 76.4 & 66.1 
& 40.7 & 82.2 & \underline{36.8} & 16.7 & 53.2 & 45.9 \\
Qwen3-VL-Embedding~\cite{li2026qwen3} 
& \textbf{94.6} & \underline{77.5} & \textbf{71.2} & \underline{67.2} & \underline{76.9} & \underline{77.5} 
& \underline{53.8} & \underline{87.2} & \textbf{56.1} & \textbf{32.8} & \underline{64.8} & \underline{58.9} \\
\midrule
TARA $\oplus$ Q3VLE & \underline{94.3} & \textbf{78.3} & \underline{70.0} & \textbf{68.6} & \textbf{81.4} & \textbf{78.5} & \textbf{54.5} & \textbf{88.4} & \textbf{56.1} & \underline{32.1} & \textbf{66.2} & \textbf{59.5} \\
\bottomrule
\end{tabular}%
}
\label{tab:mmebv2-subset}
\vspace{-5mm}
\end{table}

\begin{figure}[h!]
\vspace{-2mm}
\centering

\begin{subfigure}{\linewidth}
    \centering
    \includegraphics[width=\linewidth]{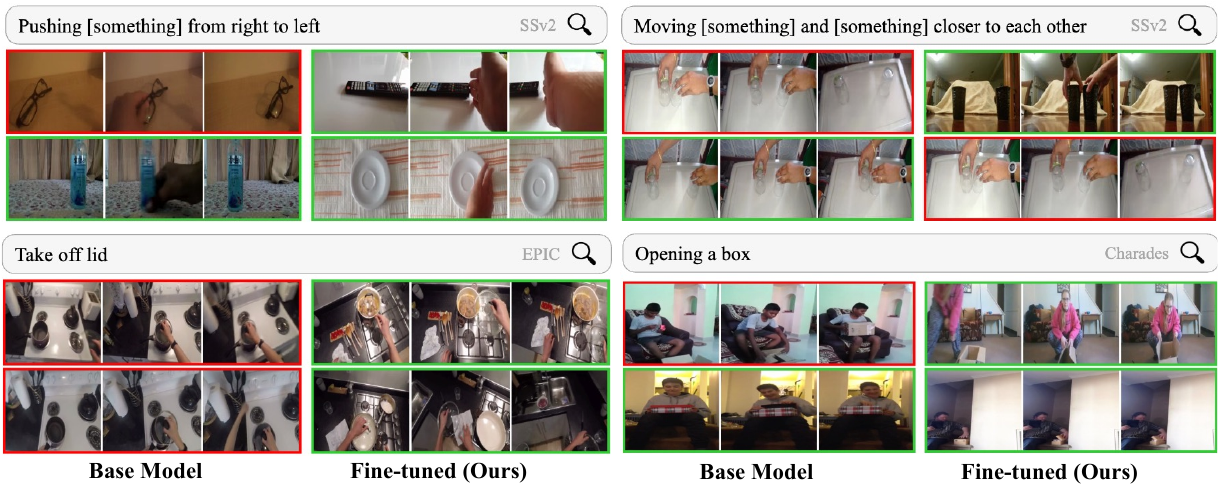}
    \caption{\scriptsize  \textbf{Chiral actions.} Queries require \textit{temporal nuance}. For each query, we show top-2 retrieved videos (green border: correct result). \TARA retrieves accurately particularly for chiral queries.
    }
    \label{fig:time-nuance}
\end{subfigure}

\begin{subfigure}{\linewidth}
    \centering
    \includegraphics[width=\linewidth]{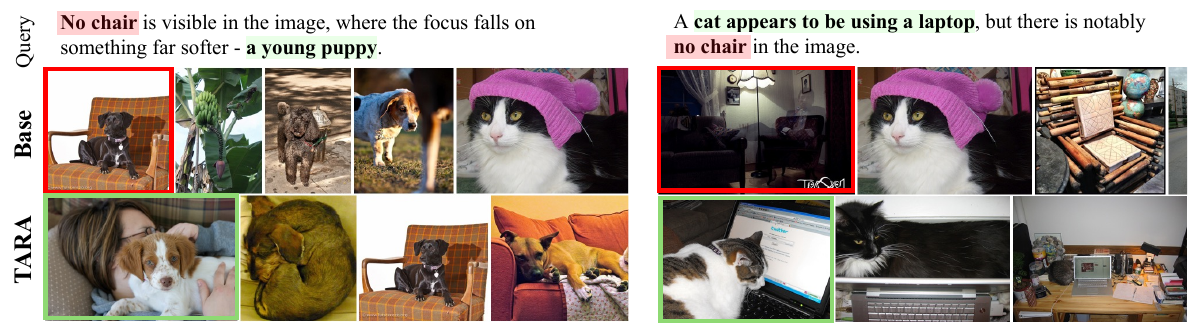}
    \caption{\scriptsize \textbf{Negation in query.} Base model is distracted by negation, while \TARA fine-tuned model retrieves accurately. Top 1 results is marked by red (incorrect) or green (correct).}
    \label{fig:language-nuance}
\end{subfigure}

\begin{subfigure}{\linewidth}
    \centering
    \includegraphics[width=\linewidth]{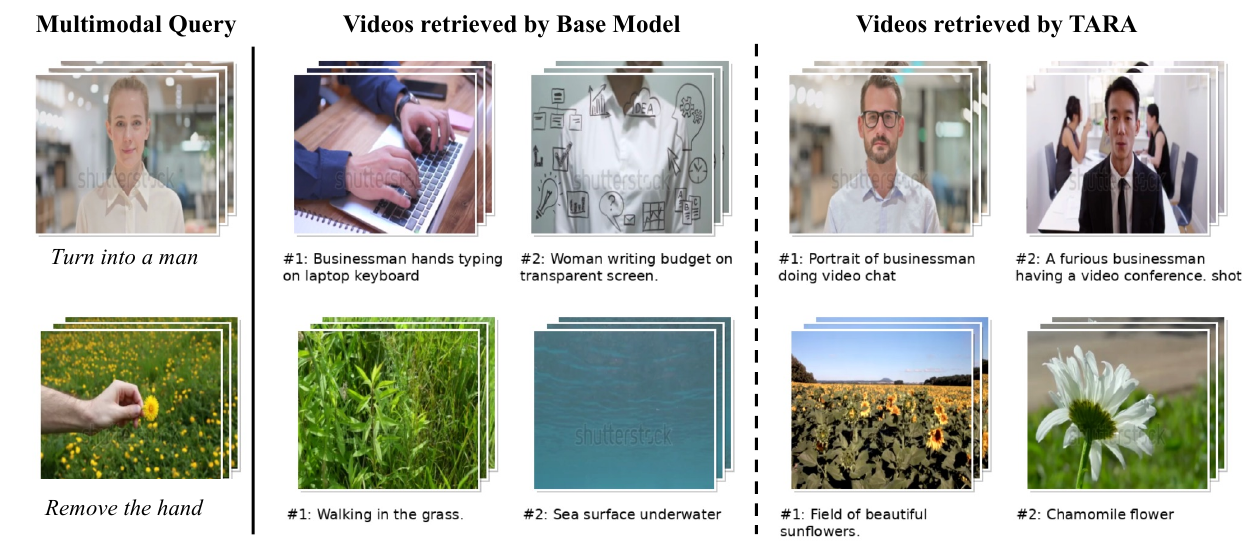}
    \caption{\scriptsize  \textbf{Composed video retrieval.} (Left) Input query composed of a video and edit instruction. Top-2 results from base model (middle) and \TARA (right). Captions of candidates are not used.}
    \label{fig:multimodal-nuance}
\end{subfigure}

\caption{Qualitative retrieval results under different types of nuance.}
\label{fig:qual-results}

\vspace{-4mm}
\end{figure}

\noindent\textbf{Qualitative results.}
We visualize qualitative retrieval results for each of three nuances: temporal nuance in \cref{fig:time-nuance}, negation nuance in \cref{fig:language-nuance} and multimodal nuance in \cref{fig:multimodal-nuance}. In each case, \TARA fine-tuned only on text, retrieves results accurately compared to the base model. 

\vspace{2mm}
\noindent\textbf{Controlled evaluation with fixed base model.} To reinforce the effectiveness of \TARA, we compare with methods that fine-tune the same base model (Qwen2VL) with multimodal data and various objectives. Results are shown in \cref{tab:qwen2vl-tasks}. On avg. across our benchmarks for nuanced retrieval, \TARA is +10 points better than the second best method despite being fine-tuned only on text.

\begin{table}[h]
\centering
\caption{\textbf{Comparison with the same base model.} We compare methods that use the same base model but fine-tune it with various data. \TARA fine-tuned on text alone almost always outperforms existing work that uses multimodal data.}
\vspace{-2mm}
\resizebox{0.9\columnwidth}{!}{%
\begin{tabular}{lccccccccc}
\toprule
\multicolumn{1}{c}{\textbf{Method}} & 
\multirow{2}{*}{\textbf{\begin{tabular}[c]{@{}c@{}}Fine-tuning \\ modalities\end{tabular}}} & 
\multicolumn{3}{c}{\textbf{CiA (SSv2)}} & 
\multicolumn{2}{c}{\textbf{RTime}} & 
\multicolumn{2}{c}{\textbf{WebVid-CoVR}} & 
\textbf{Avg.} \\
\cmidrule(lr){3-5} \cmidrule(lr){6-7} \cmidrule(lr){8-9} 
 &  & Chiral & Static & All & T2V & V2T & R@1 & R@5 & - \\
\midrule
Base (Qwen2VL-7B) & - 
& 60.2 & 28.0 & 17.3 & 59.9 & 64.7 & 15.5 & 34.4 & 40.0 \\

ArrowRL~\cite{aot-pickup2014seeing} & \faFilm \ \faPencil 
& 67.5 & 33.8 & 22.5 & 57.1 & 68.8 & 41.8 & 67.9 & 51.3 \\

CaRe (Stage 2)~\cite{xu2024fine} & \faFilm \ \faPencil 
& 66.4 & \textbf{46.2} & 23.7 & 59.8 & 69.9 & 35.6 & 61.8 & 51.9 \\

LAMRA~\cite{liu2025lamra} & \faPhoto \ \faPencil  
& 55.3 & 15.2 & 7.8 & 57.9 & 63.9 & 31.5 & 56.1 & 41.1 \\

VLM2Vec-2.0~\cite{meng2025vlm2vec} & \faFilm \ \faPhoto \ \faPencil
& 58.8 & 27.8 & 15.9 & 54.3 & 61.8 & 42.9 & 67.1 & 46.9 \\

\rowcolor{green!8}
Base + \TARA (Ours) \ & \faPencil 
& \textbf{72.7} & 39.6 & \textbf{28.3} & \textbf{65.9} & \textbf{73.8} & \textbf{44.8} & \textbf{72.6} & \textbf{56.8} \\

\bottomrule
\end{tabular}%
}
\label{tab:qwen2vl-tasks}
\vspace{-2mm}
\end{table}







\section{Analysis}
\label{sec:analysis}

\subsection{Why is text-only training effective? A study on modality gap}

\begin{figure}[h]
\vspace{-7mm}
    \centering
    \includegraphics[width=\linewidth]{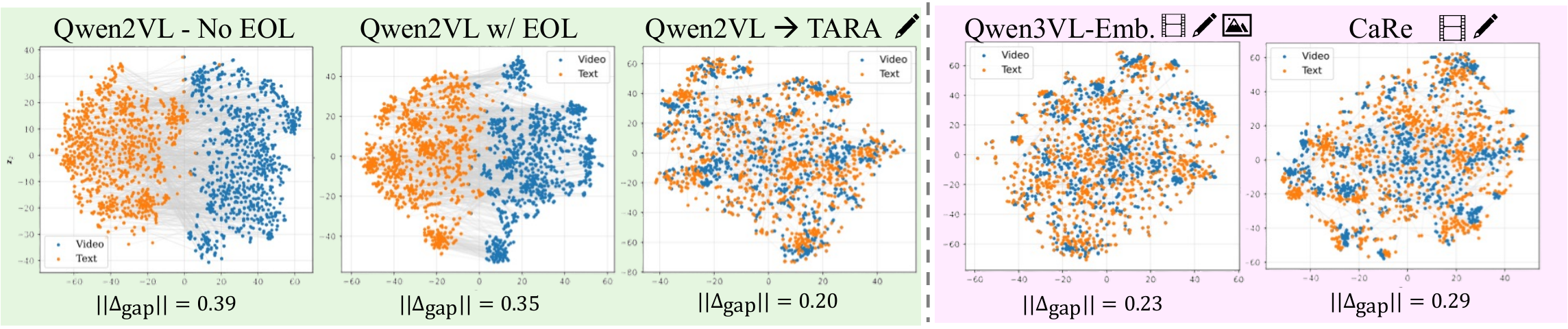}
    \caption{\textbf{Modality Gap in MLLMs} as measured on video-text pairs from MSRVTT. 
    We show results for TARA with a Qwen base model since we are comparing with others also from the Qwen family.
    (i) (\mycolorbox{green!8}{Left}) Using the EOL prompt with the Qwen2VL model alone does not fix the modality gap. We show that text-only \faPencil fine-tuning is sufficient to reduce the modality gap. (ii) 
    (\mycolorbox{pink!25}{Right}) In contrast, methods like Qwen3VL-Embedding~\cite{li2026qwen3}, CaRe~\cite{xu2024fine} train on large amounts of video \faFilm / image \faPhoto / text \faPencil data, and consequently have a reduced modality gap. 
    }
    \label{fig:modgap-mllms-v2}
\end{figure}

\noindent\textbf{Experimental setting.} The modality gap in MLLMs is relatively understudied, particularly in the context of video-text retrieval.
We use the evaluation set of MSRVTT to visualize and measure the modality gap. We pick $N{=}1000$ video-text pairs, embed them through an MLLM using a standard EOL prompt~\cite{jiang2024e5} and measure the modality gap in the shared embedding space based on definitions in~\cite{liang2022mind}.
It is denoted by ${\Delta}_{\text{gap}}$ defined as the avg. of difference between embeddings of paired samples from two modalities and we report norm of this vector.
tSNE projections and $||{\Delta}_{\text{gap}}||_{2}$ are shown in \cref{fig:modgap-mllms-v2}. 
\vspace{1mm}

\noindent\textbf{MLLMs exhibit modality gap despite shared backbone.}
Unlike dual-encoder models like CLIP, MLLMs share the LLM backbone that ingests token embeddings either from video or text. Despite that, we observe clear separation between modalities for embeddings from Qwen2VL as also measured quantitatively based on the metric from~\cite{liang2022mind} (similar results for InternVL3, Tarsier, Qwen3VL included in \cref{asubsec:mogdap}). We hypothesize that this happens because video and text tokens come from two different pathways. Video is tokenized via a pre-trained vision encoder and then projected in the LLM space by an MLP while text is tokenized via learned embedding matrices.
Modality gap hinders retrieval capabilities in several ways: (i) many components of the embeddings end up encoding the modality wasting representational capacity to embed fine-grained semantics, (ii) cross-modal cosine similarities are skewed by this gap and show smaller variation which affects discrimination between fine-grained pairs. 
\vspace{-4mm}

\noindent\textbf{EOL prompt does not reduce modality gap; text-only training does!} \mycitet{Jiang}{jiang2024e5} using LLaVA-Next-8B as the base model report that for image-text data, using the EOL prompt dissolves the modality gap. However, we find this is untrue for video-text data using Qwen2VL as shown in the first two images in \cref{fig:modgap-mllms-v2} (as well as for InternVL3, Qwen3VL, Tarsier as reported in \cref{asubsec:mogdap}).
Contemporary methods such as CaRe~\cite{xu2024fine}, Qwen3VL-Embedding~\cite{li2026qwen3} trained on large amount of video-text data show reduced modality gap. Other prior work uses various hand-crafted heuristics to reduce this gap~\cite{C3, pmlr-v280-yaras25a, MixedModalitySearch, yu2025unicorn}. In contrast, we discover that simple text-only fine-tuning is sufficient to reduce the modality gap as shown in the last image.

But why does this happen? We hypothesize two reasons for this: (i) unimodal contrastive learning leads to \textit{uniformity} pressure, i.e., it tends to spread apart the embeddings to occupy a larger subspace of the hypersphere~\cite{papyan2020prevalence,wang2020hypersphere}. This uniformity makes the text embeddings more centered around the origin. Since the LLM shares the weights while ingesting a video, the video embeddings also spread around the origin. This means the centroids of both text/video spaces move closer to the origin and thus the reduction in modality gap. 
Additionally, the MLLM
is typically pre- and post-trained to map video to text whereas here,
we use text-only contrastive training which makes a difference to the MLLM’s
embedding space
(ii) Since the negatives in the contrastive setup are also text samples, it encourages the embedding to ignore the modality information and focus more on semantics. This leaves more room for the embeddings to encode semantics thus improving retrieval capabilities. 
In \cref{aubsec:modgap-experiments}, we also empirically show (1) modality gap ($\lVert\Delta_{\text{gap}}\rVert$ from \cite{zhang2023diagnosing}) reduces in tandem with the training loss;
the downstream retrieval (SSv2) improves as $\lVert\Delta_{\text{gap}}\rVert$ decreases.
(2) We also measure the \textbf{uniformity pressure loss} ($\mathcal{L}_{\text{uniform}}$ defined in \cite{wang2020hypersphere}) in each modality. In both modalities, TARA reduces uniformity loss compared to the base model,
\ie, the embeddings are more spread apart in TARA.

\begin{figure}[h]
    \centering
    \includegraphics[width=\linewidth]{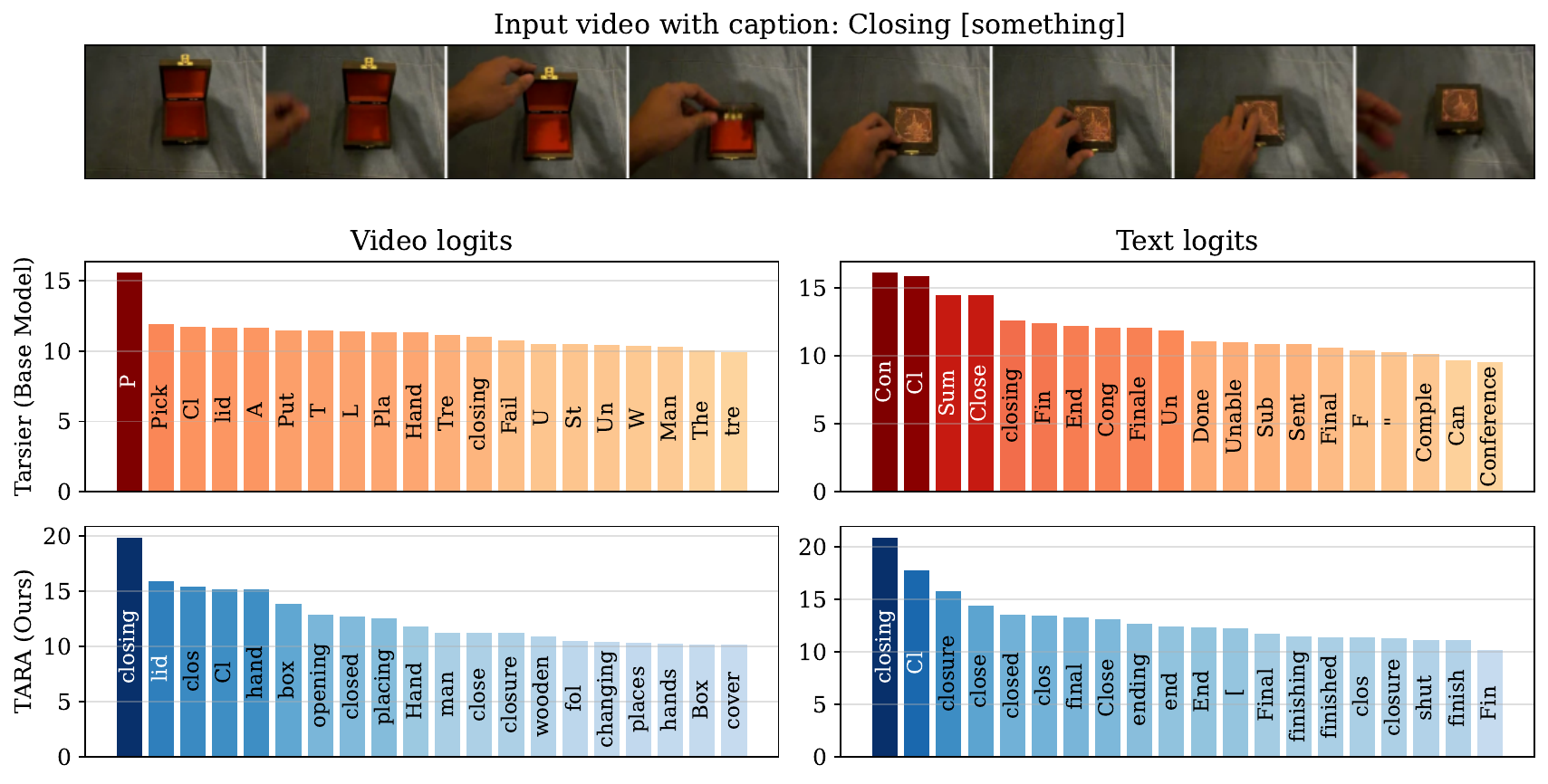}
    \caption{\footnotesize \textbf{Token distribution for video/text embedding.} Compared to base model, \TARA top tokens for the video/text correspond to semantically relevant action in the video, \ie, ``closing the box''. }
    \label{fig:token-dist}
    \vspace{-4mm}
\end{figure}

\begin{figure}[h]
    \centering
    \includegraphics[width=\linewidth]{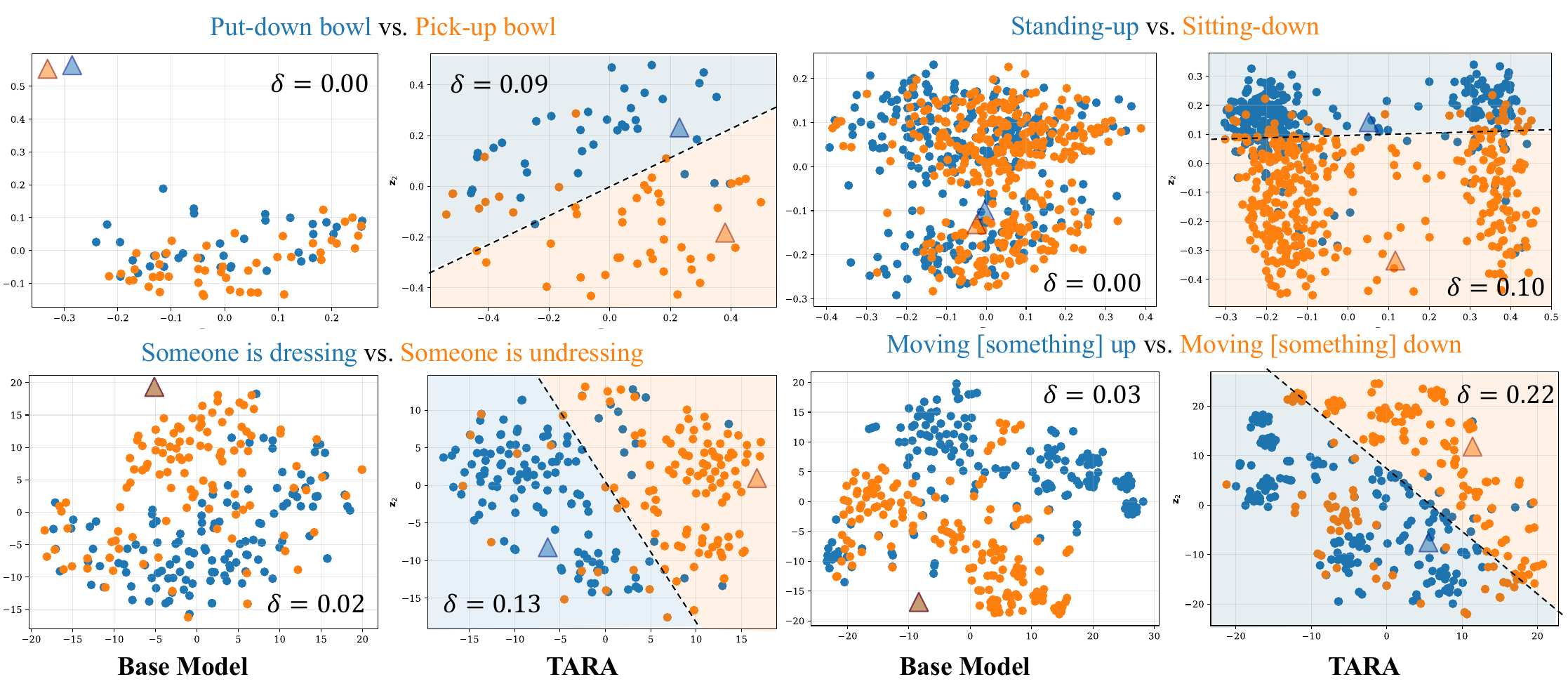}
    \caption{
    \footnotesize \textbf{Visualizing video-text embeddings.} For each chiral pair, we visualize embeddings before and after \TARA fine-tuning (text embeddings: $\triangle$). Post fine-tuning, video embeddings separate better, and text embeddings align with their corresponding videos due to reduced modality gap. We quantify this alignment ($\delta$) as the difference in avg.\ similarity between text and matched \vs mismatched videos. For the base model, $\delta {\sim} 0$, indicating poor text-video association, whereas {\tt TARA} achieves much higher $\delta$.
    }
    \label{fig:tsne-chiral}
    \vspace{-6mm}
\end{figure}

\vspace{1mm}
\noindent\textbf{Analyzing learned embeddings.}
To interpret {\tt TARA}'s embeddings, we feed them through the LM head and visualize the top 20 tokens by logits for a video and its caption independently (\cref{fig:token-dist}). Unlike the base model which generates text only and produces a semantically meaningless first token, {\tt TARA}'s embeddings reflect the relevant action (\eg, ``closing the box''), as fine-tuning trains them to discriminate between nuanced captions. More examples in \cref{sec:qual}.
We also visualize embeddings for chiral action pairs (\eg, ``put-down'' \vs ``pick-up'') \cite{bagad2025chirality} via 2D tSNE in \cref{fig:tsne-chiral}. Fine-tuning the shared LLM weights to distinguish chiral text pairs also improves video embedding separation, while the reduced $\lVert\Delta_{\text{gap}}\rVert$ aligns text-video embeddings by class, enabling strong retrieval.


\subsection{Theoretical analysis}

To answer why text-only training is so effective in cross-modal retrieval, we also theoretically analyze the two losses: $\mathcal{L}_{\text{text-only}}$ (text-only contrastive loss) and  $\mathcal{L}_{\text{t2v}}$ (text-to-video contrastive loss).
By using the characterization of modality gap in prior work~\cite{zhang2024connect,zhang2023diagnosing} (MGA), we show that 
\begin{equation}
    \mathcal{L}_{\text{text-only}} \leq \mathcal{L}_{\text{t2v}} \leq k. \mathcal{L}_{\text{text-only}}, \text{where} \ k \ \text{is a constant} > 1.
\end{equation}
This implies 
$1 \le \mathcal{L}_{\text{t2v}} / \mathcal{L}_{\text{text-only}} \le k$,
so driving the text-only loss to zero drives the text-to-video loss to
zero at the same rate up to the constant that depends on the alignment noise. This provides a theoretical basis for using our text-only loss for fine-tuning.

\begin{assumption}[Modality Gap Assumption (MGA)~\cite{zhang2024connect,zhang2023diagnosing}]
For a given video text pair $(v,t)$, the corresponding embeddings
$\mathbf{v},\mathbf{t}$ satisfy
$
\mathbf{v}-\mathbf{t}
= \mathbf{c}_{\perp}+\boldsymbol{\epsilon},
$
where $\mathbf{c}_{\perp}$ is a constant vector representing the
modality gap, orthogonal to both embedding spans, and
$\boldsymbol{\epsilon}\sim\mathcal{N}(\mathbf{0},\nu^2\mathbf{I})$
is alignment noise. See Fig.~2 of \cite{zhang2024connect}.
\end{assumption}
\begin{proposition}[Ours]
\label{prop:prop2}
Under the MGA, with unit-norm query embeddings, the text-to-video loss of
any triplet, after averaging over the alignment noise, satisfies
\begin{equation}
\ell^{\text{text-only}} \leq \ell^{t2v} \leq e^{\nu^{2}}. \ell^{\text{text-only}}.
\label{eq:bounds}
\end{equation}
This holds pointwise per triplet with triplet-independent
constants, hence also between the population losses:
$\mathcal{L}_{\text{text-only}} \leq \mathcal{L}_{t2v} $.
\end{proposition}
\begin{proof}[Proof sketch]
The proof is based on re-writing the text-to-video loss in terms of the text embeddings only and a noise term:
\[
\ell^{t2v} = \mathbb{E}_{z}\!\Big[ -\log\sigma\!\big(
\mathbf{t}_q^\top(\mathbf{t}_+ - \mathbf{t}_-) + z \big) \Big]
\]
where $z = \mathbf{t}_{q}.(\mathbf{\epsilon}_{+} - \mathbf{\epsilon}_{-}) \sim \mathcal{N}(\mathbf{0}, 2\nu^{2}\mathbf{I})$.
Then, applying the Jensen's inequality (for a convex function $f$ and random variable $X$, $f(E[X]) \leq \mathbb{E}[f(X)]$) with $f := \log$ (concave) yields the lower bound. Likewise, applying it with $f := -\log \sigma$ (convex) with careful algebraic manipulation yields the upper bound.
\end{proof}

\section{Conclusion and Discussion}
\vspace{-2mm}
In this work, we proposed a recipe, {\tt TARA}, to adapt an MLLM for various kinds of nuances in video retrieval: temporal, negation and multimodal. It is based on fine-tuning with contrastive loss on text triplets where the hard negatives are chosen carefully to inject the desired nuances in the embedding space. Our method achieves state of the art performance on all benchmarks evaluating nuanced retrieval while also improving upon the base model on standard retrieval datasets. We also present an analysis of why text-only fine-tuning is so effective through the lens of the modality gap. Specifically,  contrastive training on text triplets reduces the modality gap as a result of the uniformity pressure~\cite{wang2020hypersphere}, \ie, the video and text subspaces expand on the hypersphere and thus get closer to each other in the process. A reduced gap leads to better organization of the multimodal embedding space which in turn leads to stronger retrieval.

Since \TARA works well across different base MLLMs, it opens up the possibility of incorporating complementary advances across the LLM space (reasoning models, test-time optimization, mixture of experts, \etc). While training on text alone is a strength of {\tt TARA}, investigating ways of utilizing videos (and other modalities) in the train set to obtain a truly universal encoder remains an interesting avenue for future research. Likewise, exploring ways of using \TARA in a two-stage (retrieval-and-reranking) setup should further boost performance.

\vspace{2mm}
{\small \noindent\textbf{Acknowledgments.} This research is funded by
the EPSRC Programme Grant VisualAI EP/T028572/1, and a Royal Society Research Professorship  RSRP$\backslash$R $\backslash$241003. 
We are also grateful for funding from Toshiba Corporation.
We thank Yuta Shirakawa and Stephan Liwicki from Toshiba for comments and suggestions. We thank Ashish Thandavan for support with infrastructure. We also thank the anonymous reviewers for improving this work.}

%
%
{
\footnotesize
\bibliographystyle{splncs04}
\bibliography{main}
}

\pagebreak




\begin{center}
    {\Large\textbf{Appendix}}
\end{center}



\vspace{2mm}
\printappendixtoc   

\appendix

\asection{\textit{NLI-Nuance} Training Dataset Details}
\label{asec:nlidata}

In this section, we describe in more detail the processing used to obtain the training dataset. For example, generating temporal antonym sentences (\cref{subsec:temp-anto}), replacing anonymous subjects with realistic subjects in Ego4D captions (\cref{subsec:subject}), detecting negation in NLI triplets (\cref{subsec:negdec}), and some example triplets in the \textit{NLI-Nuance} dataset (\cref{subsec:ex-tripl}).

\asubsection{Generating (temporal) antonym sentence}
\label{subsec:temp-anto}
We prompt the Qwen3-1.7B LLM model to generate temporally antonymous sentences for Ego4D captions. We use some in-context samples to guide the model. The exact prompt with few in-context samples is provided below. Based on our initial filtering of chiral verbs, we use this stage to get antonyms for 425K sentences. 

An alternative to using an LLM for this stage is to simply detect and replace the chiral verb phrase. But we observe that in a lot of cases the antonym involves more than just replacing a chiral verb, \eg, ``pushing something from left to right'' should become ``pushing something from right to left''. Using an LLM instead of hard-coded rules ensures against errors in such cases.
\vspace{2mm}

\noindent\textbf{Quality verification.} We conduct a thorough verification of all $n{=}1000$ temporal hard negatives (THNs) generated by Qwen3 using an \textbf{LLM-as-a-judge} approach (Gemini 3.0 Flash) followed by a \textbf{manual review}. We define success criteria in two ways: (i) \textit{action verb} in THN must be temporally antonymous to that in anchor, (ii) \textit{static properties} (e.g., subject) must be unchanged. The LLM Judge deems a 98.9\% success rate on (i) and 98.5\% on (ii). Example failure cases include ``rolling dough'' vs.\ ``folding dough'' or prepositional imprecision (``puts hand into machine'' vs.\ the more natural `onto machine'). We also manually review $m = 100$ THNs from the ones labeled as success. We find all of these indeed have accurate hard negatives.

\begin{tcolorbox}[colback=gray!7, colframe=gray!70, arc=2mm, boxrule=0.5pt,title=\textbf{\small Prompt for temporal antonym generation.},coltitle=black,colbacktitle=gray!25]
{\footnotesize 

    You are a helpful assistant expert at natural language understanding and grammatical nuance.\\
    
    You are given a caption.
    Your task is to generate a temporally antonymous version of the caption. \\

    You should retain the broader context of the caption but only change the action
    described in the caption as if the video is temporally reversed. \\

    In case where the verb phrase in the caption may not have a temporal antonym,
    you should return the None as the output. Never return the same caption as the output. \\

    Here are some examples:\\
    
    (1) Caption: \#C C unrolls the yarn from her left index finger\\
        Output: \#C C rolls the yarn onto her left index finger \\

    (2) Caption: \#C C folds the cloth\\
        Output: \#C C unfolds the cloth\\

    (3) Caption: \#C C puts the pan on the stove\\
        Output: \#C C takes the pan off the stove\\

    (4) Caption: Someone is walking on the street\\
        Output: None\\

    (5) Caption: \#C C checks the cloth\\
        Output: None\\
    
    Output in a JSON format {'caption\_forward': ..., 'caption\_reverse': ...}
    where caption\_forward is the original caption and caption\_reverse is the temporally
    reversed caption.
}
\end{tcolorbox}

\asubsection{Subject replacement in Ego4D captions}
\label{subsec:subject}
The subset of NLI-Nuance with temporal nuance is composed of triplets involving a temporally opposite verb in captions mined from Ego4D~\cite{grauman2022ego4d}. However, Ego4D captions are anonymized, \eg ``\#C C" denotes camera wearer. In order to make the sentences more realistic, we replace it with a plausible subject (\eg, ``A man'') by prompting an LLM. We found that using Qwen3 does not produce sufficient diversity and it often copies the subjects from the provided in-context samples.
Thus, we used a closed-source model (Gemini 2.5 Flash Lite).
We provide one in-context example for reference.
The triplet (anchor, positive, hard-negative) are all passed together for subject replacement.
The detailed prompt is provided below.

\begin{tcolorbox}[colback=gray!7, colframe=gray!70, arc=2mm, boxrule=0.5pt,title=\textbf{\small Prompt for subject replacement.},coltitle=black,colbacktitle=gray!25]
{\footnotesize
You are an expert in English comprehension and writing.\\[6pt]
Given three sentences where the subjects may be anonymized, your task is to fill the placeholders for subjects with realistic subjects.\\[6pt]
For example, given these sentences,\\[6pt]
S1: \#C C Puts down a serving spoon and chop sticks on a cooking pot\\
S2: \#C C puts a spoon in a bowl.\\
S3: \#C C Picks up a serving spoon and chop sticks from a cooking pot\\[6pt]
a valid response could be something like:\\[6pt]
S1: The chef puts down a serving spoon and chop sticks on a cooking pot\\
S2: The chef puts a spoon in a bowl.\\
S3: The chef picks up a serving spoon and chop sticks from a cooking pot\\[6pt]
This is only an example, think logically what subject would best fit the given description and situation. For example, you will not find a cook doing carpentary. In case you are not sure, you can use generic subject pronouns like ‘The man’ or ‘The person’ or ‘The lady’, or use proper nouns like name of a person etc. Do not just use the template examples, you can be slightly creative. Make sure it is the same subject in all three sentences.\\[6pt]
Test input:}
\end{tcolorbox}

\asubsection{Negation detection}
\label{subsec:negdec}
The NLI dataset~\cite{gao2021simcse} comprises 275K triplets, each consisting of an anchor sentence, a positive example defined by an entailment relationship, and a hard-negative example derived from a contradiction pair. While contradiction and negation are related concepts, they are not synonymous: contradiction encompasses a broad range of semantic conflicts, whereas negation specifically involves the direct logical denial of a proposition stated in the anchor. To systematically identify triplets in which the hard-negative constitutes an exact \textit{negation} of the anchor, rather than a more general form of contradiction, we employ Qwen3~\cite{bai2025qwen3} as an automated annotator. Specifically, we prompt the model with a carefully designed instruction that asks it to determine whether the hard-negative is a strict negation of the anchor sentence, as opposed to a loosely related or topically divergent contradiction. Out of the full set of 275K triplets, this filtering procedure identifies 79,595 triplets in which the hard-negative qualifies as a direct negation of the anchor. We then sample from this filtered subset to construct the negation portion of our NLI-Nuance benchmark, ensuring that the resulting evaluation set specifically targets a model's ability to distinguish between semantically similar sentences that differ only by the presence or absence of negation.

\begin{tcolorbox}[
colback=gray!7,
colframe=gray!70,
arc=2mm,
boxrule=0.5pt,
title=\textbf{\small Prompt for negation detection.},
coltitle=black,
colbacktitle=gray!25
]

\footnotesize

\textbf{Task: Explicit Negation Detection in NLI Hard Negatives}

You are a linguistic analyst. Given an NLI triplet (anchor, positive, hard\_negative), determine whether the hard\_negative contradicts the anchor using \textbf{explicit negation words}.
\vspace{4mm}

\textbf{Step 1: Identify Explicit Negation Words}

Scan the hard\_negative for the following negation words or morphemes:

\{ not, n't (isn't, doesn't, won't, can't, couldn't, hasn't, weren't, etc.), no, none, never, nobody, nothing, nowhere, neither, nor \}
\vspace{4mm}

\textbf{Step 2: Determine Negation}

If at least one explicit negation word appears in the hard\_negative \textbf{and} it is used to deny a claim made in the anchor, then:

\texttt{is\_negation = true}

Otherwise:

\texttt{is\_negation = false}
\vspace{4mm}

\textbf{Critical Rules}

\begin{itemize}
\footnotesize
\item ONLY explicit negation words count.
\item Antonyms (e.g., ``smiled'' vs.\ ``frowned'') are NOT negation.
\item Replacements (e.g., ``red brick'' vs.\ ``gray marble'') are NOT negation.
\item Restrictors (``only'') are NOT negation.
\item Opposite adjectives (``slow'' vs.\ ``fast'') are NOT negation.
\item The negation word must appear in the hard\_negative sentence itself.
\item Embedded negation words still count.
\end{itemize}

\textbf{Output Format}

Respond ONLY with a JSON object (no additional text).

\end{tcolorbox}

\asubsection{More example triplets}
\label{subsec:ex-tripl}
\vspace{-1mm}

More example triplets from our NLI-Nuance dataset are provided in \cref{tab:text-triplets-suppl}. While NLI triplets focus on coarse, static aspects of video understanding, we add triplets for (i) temporal nuance, (ii) negation nuance, and (iii) multimodal nuance.

\begin{table}[t]
\centering
\caption{\textbf{More samples from NLI-Nuance.} While samples from NLI focus on static bias (corresponding videos can be distinguished by a single frame), we augment it with samples that require nuanced understanding.
Words marked in \textcolor{blue}{blue} represent synonymous parts while those in \textcolor{red}{red} represent the antonymous parts of the sentence. 
}
\footnotesize  
\begin{tabularx}{\columnwidth}{X @{\hspace{8pt}} X @{\hspace{8pt}} X}
\toprule
$\mathbf{t}$ (Anchor) & $\mathbf{t}^{+}$ (Positive) & $\mathbf{t}^{-}$ (Hard negative) \\
\midrule
\rowcolor{gray!10} \multicolumn{3}{l}{\textbf{NLI}} \\
\midrule
A group of performers \textcolor{blue}{sing} a song. & The performers are \textcolor{blue}{singing}. & The performers are \textcolor{red}{painting pictures}. \\
\arrayrulecolor{gray!50}\midrule\arrayrulecolor{black}
Man \textcolor{blue}{falling off} a blue surfboard in the ocean. & Man is \textcolor{blue}{outside}. & The man is \textcolor{red}{at church}. \\
\arrayrulecolor{gray!50}\midrule\arrayrulecolor{black}
A man \textcolor{blue}{doing a back flip} while another takes a picture. & 
A man \textcolor{blue}{doing a back flip} &
A man is \textcolor{red}{laying on a tarp} \\
\arrayrulecolor{gray!50}\midrule\arrayrulecolor{black}
Hillary \textcolor{blue}{cannot be} there! & Hillary is \textcolor{blue}{not allowed} there & Hillary \textcolor{red}{must be} there. \\
\midrule
\rowcolor{gray!10} \multicolumn{3}{l}{\textbf{Temporal nuance}} \\
\midrule
The mechanic \textcolor{blue}{closes} the tool box & The mechanic \textcolor{blue}{closes} the box & The mechanic \textcolor{red}{opens} the tool box \\
\arrayrulecolor{gray!50}\midrule\arrayrulecolor{black}
The doorman \textcolor{blue}{opens} door & The doorman \textcolor{blue}{opens} the door. & The doorman \textcolor{red}{closes} door \\
\arrayrulecolor{gray!50}\midrule\arrayrulecolor{black}
The gardener \textcolor{blue}{uproots} the \textcolor{blue}{weeds} with her hand & The gardener \textcolor{blue}{uproots} \textcolor{blue}{weeds} & The gardener \textcolor{red}{plants the weeds} with her hand \\
\arrayrulecolor{gray!50}\midrule\arrayrulecolor{black}
The carpenter \textcolor{blue}{places her left hand} on the plank &
The carpenter \textcolor{blue}{places his hand} on the table &
The carpenter \textcolor{red}{removes her left hand} from the plank \\
\arrayrulecolor{gray!50}\midrule\arrayrulecolor{black}
The person \textcolor{blue}{turns off the tap} & The person \textcolor{blue}{turns off the tap} with her right hand & The person \textcolor{red}{turns on the tap} \\
\midrule
\rowcolor{gray!10} \multicolumn{3}{l}{\textbf{Negation nuance}} \\
\midrule
\textcolor{blue}{Steel is the major hardware component} for FGD. & FGD systems \textcolor{blue}{include steel}. & \textcolor{red}{Steel is not a part} of FGD  \\
\arrayrulecolor{gray!50}\midrule\arrayrulecolor{black}
An outdoor \textcolor{blue}{vendor with a table full} of hats, dolls and jewelry. & A \textcolor{blue}{vendor with items}. & The vendor has \textcolor{red}{nothing left}. \\
\midrule
\rowcolor{gray!10} \multicolumn{3}{l}{\textbf{Multimodal nuance}} \\
\midrule
{ Source text: Show hologram word-confidentiality; Edit instruction: Add the word "synergy" to the hologram description.} & Show hologram word-synergy & Show hologram word-confidentiality \\
\arrayrulecolor{gray!50}\midrule\arrayrulecolor{black}
{Source text: Llama in a field; Edit instruction: replace llama with donkey} & Donkey in a field & Llama in a field \\
\bottomrule
\end{tabularx}
\label{tab:text-triplets-suppl}
\end{table}
\FloatBarrier

\asection{Additional Experiments}

In this Section, we present some additional ablations and experiments that augment findings of the main paper. First, we ablate over size and composition of the training dataset made of text triplets. Then, we move to modality gap analysis in MLLMs. Finally, we also show results on two novel tasks: (i) adverb understanding~\cite{doughty2022you}, (ii) spatial motion retrieval (understanding of moving $\uparrow / \downarrow / \leftarrow / \rightarrow$.

\begin{figure}[h]
    \centering
    \includegraphics[width=0.9\columnwidth]{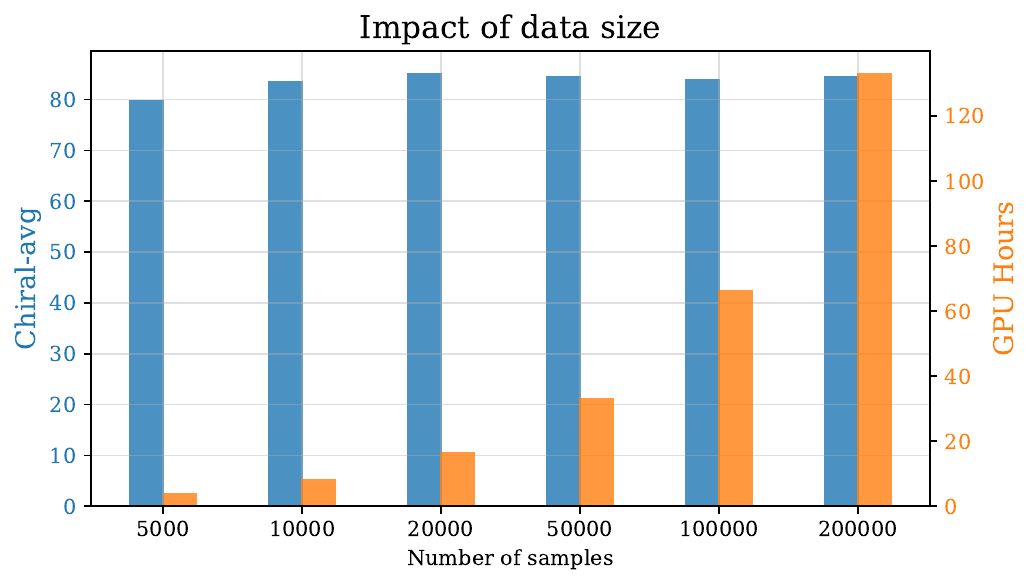}
    \caption{\textbf{Ablation on data size.} The data composition of NLI:Ego4D is fixed to 0.9:0.1 and total number of samples is varied. Using $n{=}20,000$ provides optimal results with no benefits with more text samples.
    Left scale corresponds to blue bars showing accuracy, right scale corresponds to orange bars showing GPU hours.
    }
    \label{fig:data-size}
\end{figure}

\asubsection{Ablation on data size}
\label{asubsec:data}
NLI~\cite{gao2021simcse} is made up of 275K text triplets. Since we fine-tune an MLLM which has already been pre- and post-trained on millions of samples, we question the need for such a large text dataset for contrastive training~\cite{liu2025lamra,xu2024fine}. We focus on temporal retrieval on CiA-SSv2 subset for this experiment. We only use temporal nuance triplets in training.
We fix the composition of NLI:Ego4D to be 0.9:0.1 and vary the total number of samples from $n{=}5000, \ldots, 200000$. We plot the avg.\ accuracy on the \textit{Chiral} split on SSv2 in \cref{fig:data-size}. We also plot the total GPU hours (Quadro RTX 8000) alongside the performance. The performance does not change significantly as data size increases. Thus, we pick $n{=}20,000$ as a reasonable data size that can be trained within an hour with 8 GPUs. Another reason for using low number of samples is that since we are fine-tuning the LLM weights with contrastive loss, we want to ensure that the model does not drift too far away from the base model. Training on a larger dataset reduces loss on this dataset but suffers generalization on other benchmarks.

\asubsection{Data composition}

Recall that NLI-Nuance is made up of text triplets from NLI, Ego4D (train) captions and WebVid-CoVR (train) captions. Plus, the triplets with negation also are mined from NLI. We investigate the ideal mixing ratio to build NLI-Nuance. We denote the subsets as follows: NLI (Core), Negations from NLI (Negations), Ego4D (Time) and CoVR (Multimodal).

To determine the optimal composition of our 20K-sample training set across four source datasets (Core, Negation, Time, and Multimodal), we conduct a mixture ablation study using a constrained extreme-vertices design. A naïve grid search over four mixture proportions at even a coarse 10\% resolution would require over 200 configurations, each demanding a full training and evaluation cycle, making it computationally prohibitive. Instead, we enforce a minimum Core proportion of 25\% and 5\% each for Time, Negation and systematically vary the proportions of all four sources across 22 configurations that span the feasible simplex — including boundary vertices, edge midpoints, face centroids, and interior check points. Each configuration is trained and evaluated on a fixed held-out validation set, and the results are used to fit a second-order Scheffé polynomial response surface, from which we identify the proportion allocation that maximizes downstream performance. As the validation set, we cover all three nuances: (i) Time: we use a subset of 1220 videos from the SSv2 validation set. (ii) Negation: ~\cite{alhamoud2025vision-negbench} carefully generate negative captions for images in COCO and videos in MSRVTT. Since we did not want to generate new negative captions for another dataset, we stick to the same MSRVTT set to measure negation understanding. (iii) Multimodal: We use a subset of 1000 pairs from the validation set from WebVid-CoVR~\cite{ventura2024covr}. We measure the average of metrics across the three nuances (\eg, mAP for $t{\rightarrow}v$ for chiral split of SSv2, R@1 for standard/negation on MSRVTT, R@1 for WebVid-CoVR). The overall metric is the average of these three.

\begin{figure}[h]
    \centering
    \includegraphics[width=\linewidth]{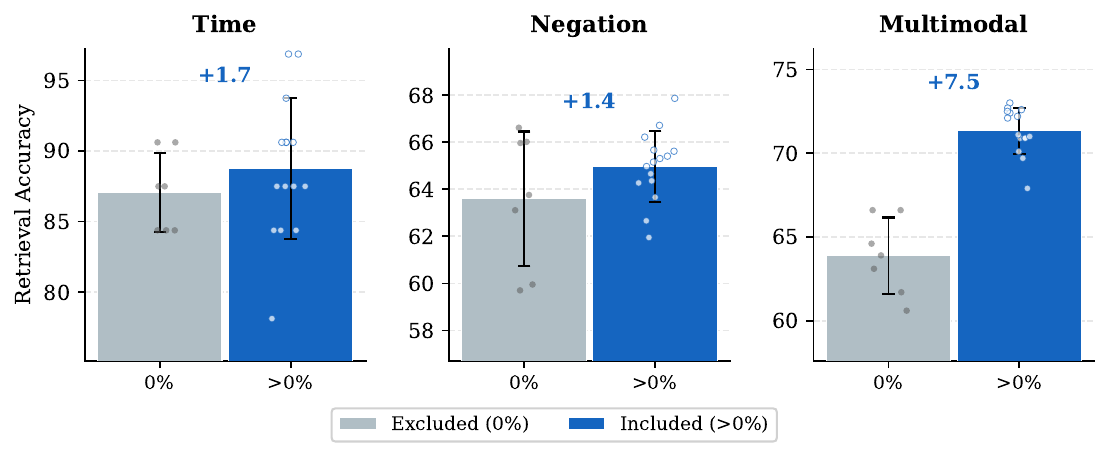}
    \caption{\textbf{Text data for each nuance matters.} We average performance on each nuance’s split with or without using text samples from that split. We confirm that using each of them is beneficial.}
    \label{fig:each-nuance}
\end{figure}

\begin{figure}[h]
    \centering
    \includegraphics[width=\linewidth]{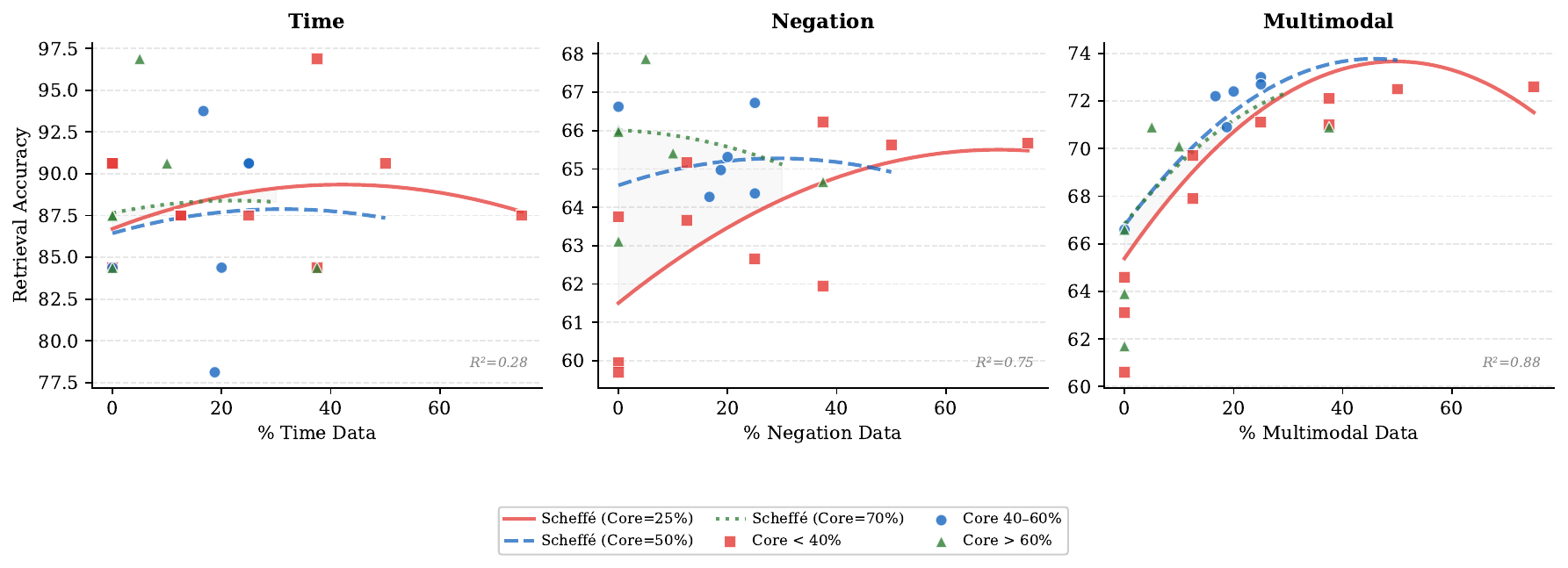}
    \caption{\textbf{Optimal data composition.}
    Red curve shows Scheffe's' fit for text samples from Core split 25\%, blue curve with Core 40-50\% and Green curve shows Core at 70\%. Starting from Multimodal split, we need to pick 50-60\% of it irrespective of Core \%. Then, looking at negation split, we need to pick Core samples at either 40-60\%. 
    As the final pick, we use (40\%, 5\%, 5\%, 50\%) as the data composition.}
    \label{fig:optimal-data}
\end{figure}

First, we confirm that using text samples for each nuance is beneficial for the corresponding nuance in \cref{fig:each-nuance}. For each nuance, we compare  its metric (\eg, SSv2 $t{\rightarrow}v$ for Time) across runs with or without using any data for that nuance. On average, including composed text retrieval data yields the largest benefit to Composed Video Retrieval. For time, the variance is higher indicating a need for careful balance of Core NLI samples and time samples. Second, for optimal data composition, we find that several runs plateau around near-best results with 40-50\% of Core samples, 5\% for Time/Negation and 40-50\% of Multimodal samples. We pick (40\%, 5\%, 5\%, 50\%) as the data composition. See \cref{fig:optimal-data} for a scatter plot for each of the three nuanced aspects.

\begin{figure}[h]
    \centering
    \includegraphics[width=\linewidth]{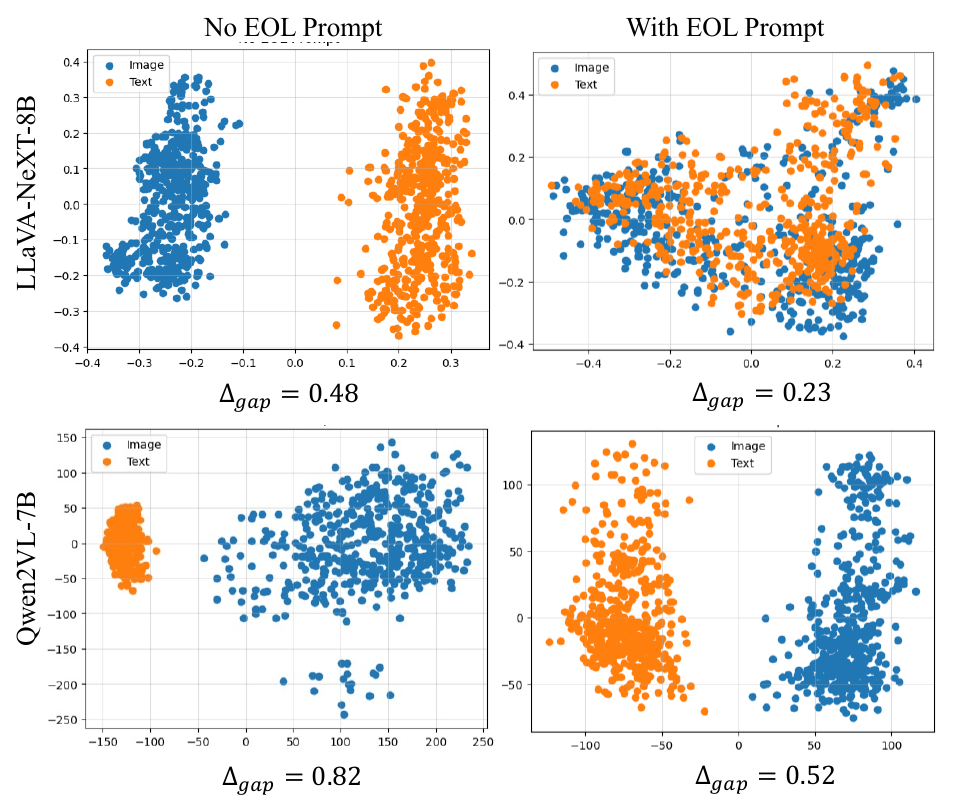}
    \caption{\textbf{Effect of EOL on LLaVA \vs Qwen.} On COCO images, we reproduced~\cite{jiang2024e5}'s finding that the EOL prompt indeed substantially reduces the modality gap for the LLaVA-NeXT-8B base model without any training.
    For Qwen2VL as well, the modality gap reduces but it starts from a higher value so even with reduced gap, the modalities are still separated. For video-text data, the reduction in gap with EOL prompt is not as substantial.
    Here, we visualize the PCA projections of the image/text embeddings. We also show quantitative measure of modality gap $||\Delta_{\text{gap}}||_{2}$.}
    \label{fig:llava-vs-qwen}
\end{figure}

\asubsection{Modality gap in other MLLMs}
\label{asubsec:mogdap}
\noindent\textbf{Reproducing finding from~\cite{jiang2024e5}.} 
\mycitet{Jiang}{jiang2024e5} find that using the EOL prompt (\eg, for an image, \texttt{[image] Summarize the image in one word: }) with the LLaVA-NeXT-8B model reduces the modality gap between image and text embeddings for image-text pairs in COCO. First, we reproduce their finding using a sample of $N{=}500$ image-text pairs from COCO (\cref{fig:llava-vs-qwen} top row). However, interestingly for Qwen2VL, using the same data, we find that the modality gap remains even with the EOL prompt (\cref{fig:llava-vs-qwen} bottom row). This shows that the modality gap also depends on the base model used.
\vspace{2mm}

\noindent\textbf{Other MLLMs.}
Since we care about video-text embeddings, as in the main paper, we use the evaluation set of MSRVTT to visualize and measure the modality gap.
In the main paper, we have already shown that Qwen2VL exhibits a modality gap between video and text embeddings with or without the EOL prompt.
Given this observation, we want to test if the EOL prompt is inadequate in dissolving the modality gap for other MLLMs: Tarsier, InternVL3, Qwen3VL, \etc. We also test LLaVA-NeXT-8B on the same MSRVTT samples to rule out the possibility that modality gap is dataset-dependent. Since it can only take in images, we feed it the middle frame of each in MSRVTT.

\begin{figure}[h]
    \centering
    \includegraphics[width=\linewidth]{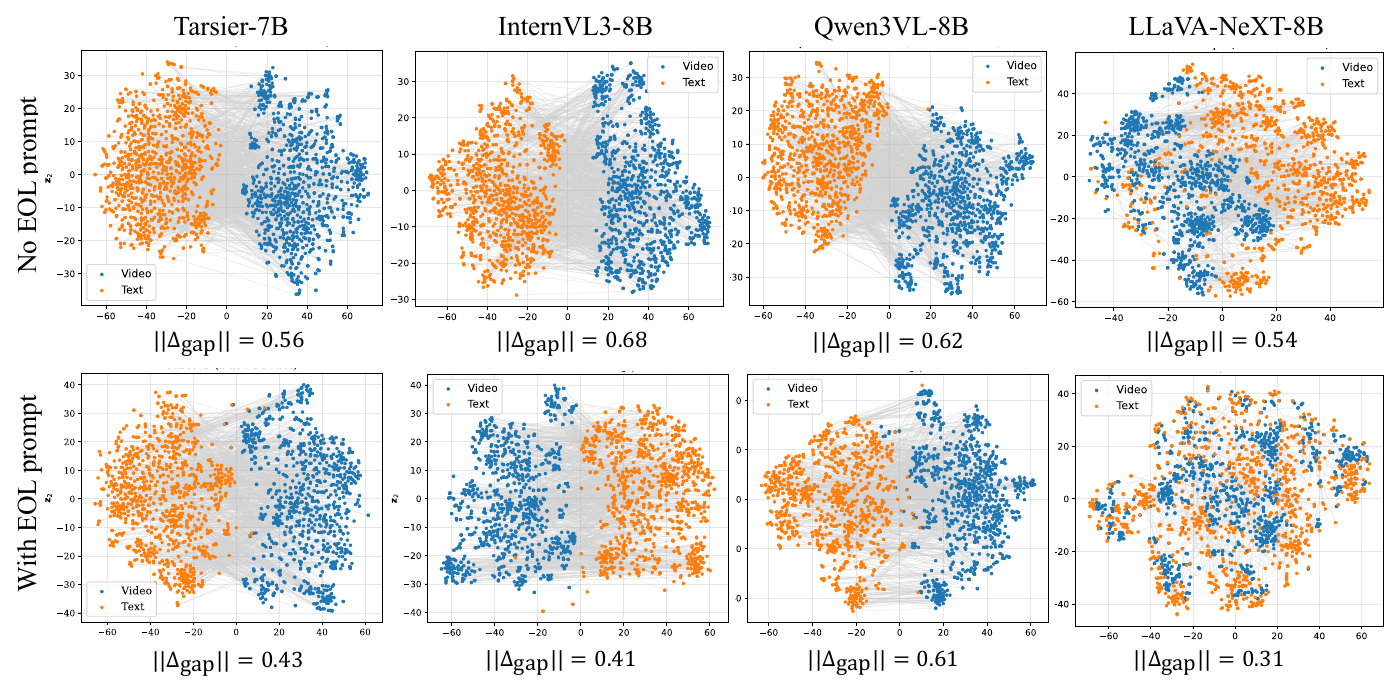}
    \caption{\textbf{Modality gap exists in other MLLMs.} 
    We visualize the tSNE projection of video-text embeddings on MSRVTT for various models.
    Like Qwen2VL, we find that the modality gap still exists for InternVL3, Tarsier and Qwen3VL even while using the EOL prompt. 
    We confirm~\cite{jiang2024e5}'s observation that the modality gap does reduce with the EOL prompt for LLaVA-NeXT even on MSRVTT samples (using the middle frame).
    }
    \label{fig:modgap-other}
\end{figure}

As shown in \cref{fig:modgap-other}, we find that much like Qwen2VL, other MLLMs like Tarsier, InternVL3, Qwen3VL all exhibit a substantial modality gap even while using the EOL prompt. However, much like on COCO images, LLaVA-NeXT does show a significant reduction in the modality gap on MSRVTT samples. This confirms that there is something about the model architecture/weights itself in LLaVA-NeXT that renders this property. Note, as we show in the main paper, text-only fine-tuning on \TARA is sufficient to reduce modality gap
for these other MLLMs.
\vspace{2mm}

\asubsection{Adverb Understanding}
\label{asubsec:adverbs}
Given a video clip $\mathbf{x} \in \mathcal{X}$ and a label for the action of interest $a \in A$, the goal of adverb recognition is to correctly predict the adverb $\hat{m}$ which applies to action $a$. For this task, we use the benchmark proposed in~\mycitet{Doughty}{doughty2022you}. Since adverbs co-occur with verbs, the task is, given a video and an action verb, select the correct adverb between two choices (the correct adverb and its antonym). For example, given a video of a person ``walking'', one needs to select if the walking is ``slow/fast''.
We encode the video together with the action verb in a single prompt.
The full prompt is provided below.
\begin{tcolorbox}[
colback=gray!3!white,
colframe=gray!85!black,
boxsep=0.7pt,
width=\linewidth,
boxrule=0.5pt
]
\footnotesize{
\texttt{USER: [video] \\
Action: This video shows the action [sent]\\
Look at the video carefully. Summarize the action in the video in one word:
ASSISTANT:
}
}
\end{tcolorbox}
Likewise, for text, we represent the verb-adverb by including them together in a sentence.
We embed the sentence description of the action with either of the two adverbs: {\small \texttt{The action [action] is performed [adverb].}} Finally, the similarity is computed in the common embedding space.

We follow the test splits for MSRVTT and VATEX in ~\mycitet{Doughty}{doughty2022you}. MSRVTT has 1,824 clips with 18 adverb pairs while VATEX has 2,835 clips with 34 adverb pairs. As baselines, we compare with (i) Action Modifiers~\cite{doughty2020action} trained on adverb subsets in HowTo100M (HTM) and VATEX (VTX), (ii) follow-up work~\cite{doughty2022you} that uses pseudo-labeling in addition to a small amount of labelled data, and (iii) embeddings from the base model Tarsier 2. Results are shown in \cref{tab:adverbs}. TARA zero-shot exceeds the semi-supervised model in~\cite{doughty2022you}.

\begin{table}[h]
\centering
\caption{\textbf{Adverb recognition.}
We compare with Action Modifiers~\cite{doughty2020action} trained on adverb subsets in HowTo100M (HTM) and VATEX (VTX) and follow-up work~\cite{doughty2022you} that uses pseudo labels along with small amount of labeled data (20\%). Psuedo labels come from their own partially trained model. \faFilm, \faPencil denote the video, text modalities in the fine-tuning data.
H2M denotes adverb subset of HowTo100M, VTX that of VATEX. 
TARA exceeds semi-supervised models trained specially on adverb recognition.
}
\resizebox{\columnwidth}{!}{%
\begin{tabular}{llcc}
\toprule
\textbf{Method} & \textbf{Fine-tuning dataset(s)} & \textbf{VATEX} & \textbf{MSRVTT} \\ \midrule
Chance & - & 50.0 & 50.0 \\
\midrule
Action Modifiers~\cite{doughty2020action} & H2M{+}VTX(20\%) (\faFilm, \faPencil) & 64.2 & - \\
AM with pseudo-labels~\cite{doughty2022you} & H2M{+}VTX(20\%) (\faFilm, \faPencil) & 67.5 & 65.0 \\
AM with pseudo-labels~\cite{doughty2022you} & H2M{+}VTX(20\%){+}MSRVTT (\faFilm, \faPencil) & 67.5 & 70.5 \\
Tarsier 2 & - & 57. 4 & 56.6 \\
\rowcolor{green!8} Tarsier 2 + TARA & NLI-Nuance (\faPencil) & \textbf{74.8} & \textbf{76.8} \\ 
\bottomrule
\end{tabular}%
}
\label{tab:adverbs}
\end{table}

\begin{figure}[h]
    \centering
    \includegraphics[width=\linewidth]{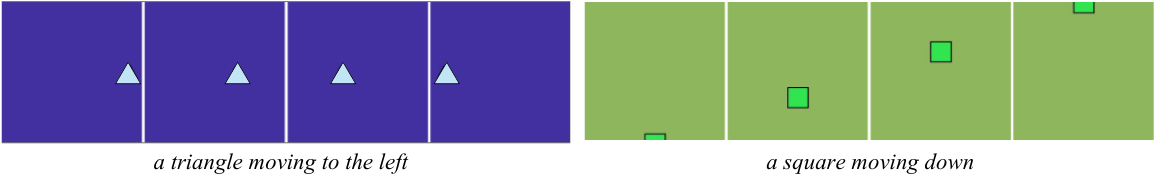}
    \caption{\textbf{Spatial motion retrieval.} Example synthetic videos to probe understanding of simple spatially moving objects.}
    \label{fig:spatial-motion}
\end{figure}

\asubsection{Spatial motion retrieval}
\label{asubsec:spatial}
We evaluate if the model embeddings encode simple spatial motion patterns such as moving something $\uparrow / \downarrow / \leftarrow / \rightarrow$. Specifically, we simulate videos of a simple-shaped object moving in one of these four directions captioned with an LLM. An example is shown in \cref{fig:spatial-motion}. Given a text query (\eg, ``a green triangle moving left''), the task is to rank the videos by embedding similarity. We measure R@1 across three attributes: (a) object shape (does the retrieved video have a triangle?), (b) motion direction (does the retrieved video feature a shape moving leftward?), and (c) both.
For each motion direction, we randomise the shape, background attributes and simulate $n{=}100$ videos amounting to a total of $400$ videos.
We show the results in in \cref{tab:results}: across all three, we confirm TARA's superiority over Qwen3VL-Embedding.

\begin{table}[h]
\centering
\caption{R@1 on synthetic spatial-motion retrieval across criteria.}
\label{tab:results}
\resizebox{\columnwidth}{!}{%
\begin{tabular}{l|c|c|c|c>{\columncolor{green!8}}c}
\toprule
\textbf{Criterion} & \textbf{Chance} & \textbf{Qwen3VL-Emb.\ 8B} & \textbf{Tarsier2 7B} & \textbf{Tarsier2 7B + TARA} \\
\midrule
Object shape     & 20.0 & 91.3          & 64.0 & \textbf{92.0} \\
Motion direction & 25.0 & 85.7          & 45.7 & \textbf{92.4} \\
Both             & 5.0  & 78.0          & 25.8 & \textbf{84.5} \\
\bottomrule
\end{tabular}%
}
\end{table}

\asubsection{Concatenation of Embeddings (CoE)}
\label{asec:moe}

\begin{figure}[h]
    \centering
    \begin{minipage}{0.55\linewidth}
        \includegraphics[width=\linewidth]{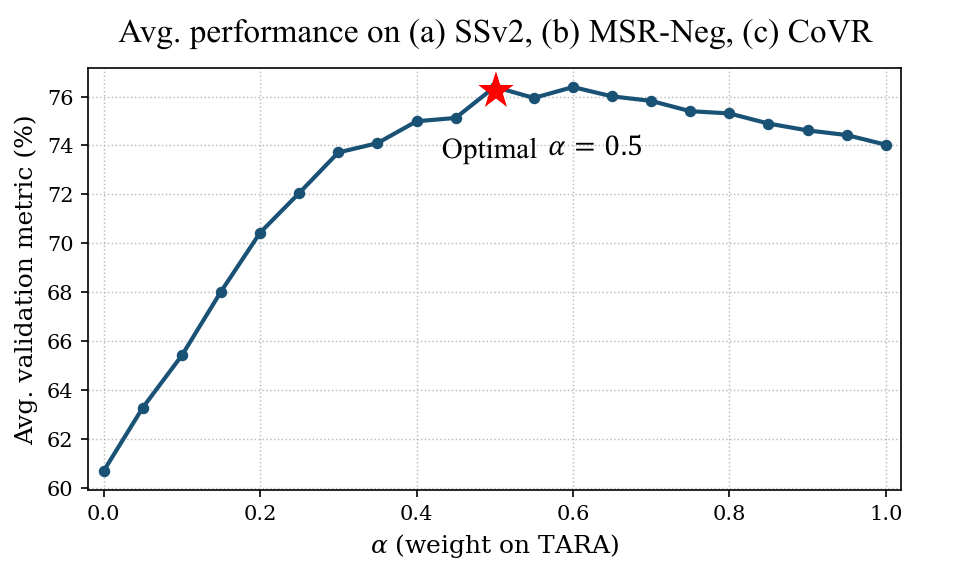}
    \end{minipage}
    \begin{minipage}{0.4\linewidth}
        \captionof{figure}{\textbf{Line search for $\alpha$.} We conduct a line search over possible $\alpha \in [0, 1]$ values. We measure the average performance on the validation sets of (a) SSv2 (CiA) R@1, (b) MSRVTT-NegBench R@5 and (c) CoVR R@1.}
        \label{fig:line-search}
    \end{minipage}
\end{figure}

We extend the experiments with concatenation of TARA and Qwen3-VL embeddings on nuanced retrieval tasks.
Here, we use $\alpha \in [0, 1]$ to denote the weight assigned to TARA's embedding in the concatenation. Concretely, let $\mathbf{z}_{a}$ denote the TARA embedding and $\mathbf{z}_{b}$ denote the Qwen3VL-Embedding. Then, $\mathbf{z} := [\alpha \mathbf{z}_{a}; (1-\alpha) \mathbf{z}_{b}]$ where $;$ denotes concatenation. Then, the similarity for a video-text pair $(v, t)$ is given by:  
\begin{align*}
    s_{\text{CoE}}(v, t) &= \operatorname{cos-sim}(\mathbf{z}^{(v)}, \mathbf{z}^{(t)}) \\
    &= \operatorname{cos-sim} \left( [\alpha \mathbf{z}_{a}^{(v)}; (1-\alpha) \mathbf{z}_{b}^{(v)}],  [\alpha \mathbf{z}_{a}^{(t)}; (1-\alpha) \mathbf{z}_{b}^{(t)}] \right) \\
    &= \alpha.\operatorname{cos-sim}\left(\mathbf{z}_{a}^{(v)}, \mathbf{z}_{a}^{(t)}\right) + (1 - \alpha).\operatorname{cos-sim}\left(\mathbf{z}_{b}^{(v)}, \mathbf{z}_{b}^{(t)}\right) \\
    &= \alpha . s_{a}(v, t) + (1 - \alpha) . s_{b}(v, t).
\end{align*}
Since nuanced retrieval consists of various datasets, we first conduct a line search over possible $\alpha$ values. We measure the average performance on the validation sets of (a) SSv2 (chiral), (b) MSRVTT-NegBench and (c) CoVR. As shown in \cref{fig:line-search}, we find $\alpha=0.5$ as the optimal choice. This is reasonably since we have a mix of fine-grained (SSv2) as well as coarse-grained (MSRVTT) datasets. Based on this choice, we evaluate on the test sets of all nuanced retrieval tasks in \cref{tab:nuanced-all-results}. We compare this ensembled model with our TARA (on Tarsier 2), the base Qwen3-VL-Embedding (Q3VL-E) model, as well as further fine-tuning Q3VL-E with the TARA text-only recipe (both full and LoRA tuning). While on temporal nuance tasks, TARA dominates Q3VL-E, on negation (MSR), Q3VL-E clearly outperforms TARA and is competitive on multimodal nuance (CoVR). However, the ensemble dominates both on almost all the tasks and is substantially better than all others on average.

\definecolor{timecol}{RGB}{220, 235, 252}
\definecolor{negcol}{RGB}{255, 230, 220}
\definecolor{mmcol}{RGB}{220, 245, 220}

\begin{table}[]
\centering
\caption{Comprehensive comparison across temporal (CiA), negation (NegBench), and multimodal benchmarks. Best-performing results in each column are shown in \textbf{bold}, while second-best results are \underline{underlined}. The final column reports the average across all metrics. Q3VL-E denotes Qwen3-VL-Embedding~\cite{li2026qwen3} and $\rightarrow$ TARA denotes further fine-tuning Q3VL-E with TARA recipe. TARA $\oplus$ Q3VL-E denotes an ensemble, \textit{concatenation of embedding}, which outperforms all other methods substantially on average.
}
\label{tab:nuanced-all-results}
\resizebox{\columnwidth}{!}{%
\begin{tabular}{@{}l *{16}{c} @{}}
\toprule[\heavyrulewidth]
\rule{0pt}{2.4ex}%
& \multicolumn{9}{c}{\cellcolor{timecol}\textbf{Time (CiA)}}
& \multicolumn{4}{c}{\cellcolor{negcol}\textbf{Negation (NegBench)}}
& \multicolumn{2}{c}{\cellcolor{mmcol}\textbf{Multimodal}}
& \textbf{Avg.} \\
\rule{0pt}{2.4ex}%
& \multicolumn{3}{c}{\cellcolor{timecol!50}\textbf{SSv2}}
& \multicolumn{3}{c}{\cellcolor{timecol!50}\textbf{EPIC}}
& \multicolumn{3}{c}{\cellcolor{timecol!50}\textbf{Charades}}
& \multicolumn{2}{c}{\cellcolor{negcol!50}\textbf{COCO}}
& \multicolumn{2}{c}{\cellcolor{negcol!50}\textbf{MSR}}
& \multicolumn{2}{c}{\cellcolor{mmcol!50}\textbf{\small WebVid-CoVR}}
& \\
\cmidrule(lr){2-4} \cmidrule(lr){5-7} \cmidrule(lr){8-10}
\cmidrule(lr){11-12} \cmidrule(lr){13-14} \cmidrule(lr){15-16}
\rule{0pt}{2.4ex}%
\textbf{Model}
& \textit{Chiral} & \textit{Static} & \textit{All}
& \textit{Chiral} & \textit{Static} & \textit{All}
& \textit{Chiral} & \textit{Static} & \textit{All}
& \textit{Std.} & \textit{Neg.}
& \textit{Std.} & \textit{Neg.}
& \textit{R@1} & \textit{R@5}
& \\
\midrule
Chance & 50.0 & 6.3 & 3.1 & 50.0 & 1.5 & 0.8 & 50.0 & 3.6 & 1.8 & 0.1 & 0.1 & 0.5 & 0.5 & 0.04 & 0.2 & 11.3 \\
Q3VL-E~\cite{li2026qwen3} & 72.0 & 43.4 & 31.8 & 62.1 & 28.6 & 20.6 & 65.3 & 37.3 & 26.1 & 78.8 & 71.8 & \textbf{79.8} & \underline{73.5} & 66.8 & 87.5 & 56.4 \\
Q3VL-E $\rightarrow$ TARA (Full) & 73.0 & 51.5 & 37.8 & 64.0 & 25.1 & 20.1 & 68.3 & 36.4 & 28.7 & \underline{80.7} & \underline{75.9} & 70.2 & 67.4 & \underline{67.6} & \underline{88.1} & 57.1 \\
Q3VL-E $\rightarrow$ TARA (LoRA) & 70.5 & 55.8 & 36.7 & 62.7 & 29.0 & 21.6 & 66.6 & 35.3 & 25.5 & \underline{80.7} & 74.9 & \underline{78.8} & 73.4 & 65.2 & 86.9 & 57.0 \\
Tarsier2 + TARA & \textbf{88.9} & \underline{66.7} & \underline{58.6} & \textbf{81.1} & \textbf{45.6} & \textbf{38.9} & \textbf{71.4} & \underline{38.6} & \underline{29.0} & 76.7 & 73.6 & 65.1 & 65.0 & 66.3 & 86.7 & \underline{60.7} \\
\midrule
TARA $\oplus$ Q3VLE & \underline{87.4} & \textbf{78.1} & \textbf{62.9} &
\underline{79.4} & \underline{43.1} & \underline{36.6} & 
\underline{71.2} & \textbf{39.4} & \textbf{29.3} &
\textbf{81.8} & \textbf{78.1} &
76.7 & \textbf{73.7} & 
\textbf{70.1} & \textbf{90.1} & \textbf{67.9} \\
\bottomrule[\heavyrulewidth]
\end{tabular}%
}
\end{table}

\asection{Theoretical Analysis}
\label{asec:theory}
In addition to the Proof Sketch provided in the main paper, we provide a more detailed proof on the connection between modality gap and the effectiveness of text-only training. For completeness, we also re-state the Propositions.

\asubsection{Detailed proofs}

\begin{assumption}[Modality Gap Assumption (MGA)~\cite{zhang2024connect,zhang2023diagnosing}]
For a given video text pair $(v,t)$, the corresponding embeddings
$\mathbf{v},\mathbf{t}$ satisfy
$
\mathbf{v}-\mathbf{t}
= \mathbf{c}_{\perp}+\boldsymbol{\epsilon},
$
where $\mathbf{c}_{\perp}$ is a constant vector representing the
modality gap, orthogonal to both embedding spans, and
$\boldsymbol{\epsilon}\sim\mathcal{N}(\mathbf{0},\nu^2\mathbf{I})$
is alignment noise. See Fig.~2 of \cite{zhang2024connect}.
\end{assumption}

\begin{tcolorbox}[
    colback=Yellow!10,
    colframe=black,
    boxrule=0.8pt,
    arc=2pt,
    left=6pt, right=6pt, top=6pt, bottom=6pt
]
\textbf{Key takeaway.} 
A pre-trained video-text embedding model can be fine-tuned with a text-only contrastive loss ($\mathcal{L}_{\text{text-only}}$) or a text-to-video contrastive loss ($\mathcal{L}_{\text{t2v}}$). We show that given the modality gap assumption, the following inequality holds under mild assumptions on the alignment noise:
\[
\mathcal{L}_{\text{text-only}} \;\leq\; \mathcal{L}_{\text{t2v}} \;\leq\; e^{\nu^2}\, \mathcal{L}_{\text{text-only}}.
\]
This implies 
$1 \le \mathcal{L}_{\text{t2v}} / \mathcal{L}_{\text{text-only}} \le e^{\nu^2}$,
so driving the text-only loss to zero drives the text-to-video loss to
zero at the same rate up to the constant $e^{\nu^2}$. This provides a theoretical basis for using our text-only loss for fine-tuning.
\end{tcolorbox}

\noindent\textbf{Setup. } Before we state the propositions and their proofs, we detail the basic notation and setup.
Consider a text query $\mathbf{t}_q$. Let $\{\mathbf{t}_+, \mathbf{t}_-\}$
be the text positive and negative for $\mathbf{t}_q$, and let
$\{\mathbf{v}_+, \mathbf{v}_-\}$ be the videos corresponding to them. All
of these are embeddings on the unit sphere in $\mathbb{R}^d$; in
particular $\|\mathbf{t}_q\| = 1$. For this triplet, the per-triplet
losses are
\begin{align}
\ell^{\text{text}} &= -\log \sigma\!\big(\mathbf{t}_q^\top(\mathbf{t}_+ - \mathbf{t}_-)\big), \\
\ell^{t2v}  &= -\log \sigma\!\big(\mathbf{t}_q^\top(\mathbf{v}_+ - \mathbf{v}_-)\big),
\end{align}
where $\sigma$ denotes the sigmoid. Under the Modality Gap Assumption
(MGA), each video--text pair $(v,t)$ satisfies
$v = t + \mathbf{c}_\perp + \boldsymbol{\epsilon}$ with
$\boldsymbol{\epsilon} \sim \mathcal{N}(\mathbf{0}, \nu^2 \mathbf{I})$, where
$\mathbf{c}_\perp$ is a constant vector (the modality gap) orthogonal to
both embedding spans, and the alignment noise $\boldsymbol{\epsilon}$ is drawn
independently across pairs.

\begin{proposition}[Lower bound]
\label{prop:lower-bound}
Under the MGA, with unit-norm query embeddings, the text-to-video loss of
any triplet, after averaging over the alignment noise, satisfies
\begin{equation}
\ell^{\text{text-only}} \leq \ell^{t2v}.
\label{eq:lower}
\end{equation}
This holds pointwise per triplet with triplet-independent
constants, hence also between the population losses:
$\mathcal{L}_{\text{text-only}} \leq \mathcal{L}_{t2v} $.
\end{proposition}

\begin{proof}[Proof of Proposition~\ref{prop:lower-bound}]
We provide a step-by-step proof as follows. \\[2mm]
\textbf{Step 1: The video margin equals the text margin plus scalar
noise.} By the MGA,
$\mathbf{v}_+ = \mathbf{t}_+ + \mathbf{c}_\perp + \boldsymbol{\epsilon}_+$ and
$\mathbf{v}_- = \mathbf{t}_- + \mathbf{c}_\perp + \boldsymbol{\epsilon}_-$, with
$\boldsymbol{\epsilon}_+, \boldsymbol{\epsilon}_- \sim \mathcal{N}(\mathbf{0},
\nu^2\mathbf{I})$ independent. Subtracting, the constant modality gap
$\mathbf{c}_\perp$ cancels:
\[
\mathbf{v}_+ - \mathbf{v}_-
= (\mathbf{t}_+ - \mathbf{t}_-) + (\boldsymbol{\epsilon}_+ - \boldsymbol{\epsilon}_-).
\]
Taking the inner product with the query,
\begin{equation}
\mathbf{t}_q^\top(\mathbf{v}_+ - \mathbf{v}_-)
= \mathbf{t}_q^\top(\mathbf{t}_+ - \mathbf{t}_-)
+ \mathbf{t}_q^\top(\boldsymbol{\epsilon}_+ - \boldsymbol{\epsilon}_-).
\label{eq:margin-split}
\end{equation}
The first term is exactly the text-only margin; the second is a scalar
random variable. \\[2mm]
\textbf{Step 2: The scalar noise is Gaussian with variance $2\nu^2$.}
Define $z := \mathbf{t}_q^\top(\boldsymbol{\epsilon}_+ - \boldsymbol{\epsilon}_-)$. Since
$\boldsymbol{\epsilon}_+$ and $\boldsymbol{\epsilon}_-$ are independent
$\mathcal{N}(\mathbf{0},\nu^2\mathbf{I})$, their difference is
$\mathcal{N}(\mathbf{0}, 2\nu^2\mathbf{I})$. 
This a weighted sum of Gaussians and thus itself a Gaussian with mean $0$ and variance $2\nu^2$.
Formally, projecting a centered
isotropic Gaussian onto the fixed unit vector $\mathbf{t}_q$ (here
$\|\mathbf{t}_q\| = 1$ is used) gives
\begin{equation}
z \sim \mathcal{N}(0,\, 2\nu^2).
\label{eq:z-law}
\end{equation}
Thus the text-to-video loss for this triplet, after averaging over the
noise, is
\begin{equation}
\ell^{t2v}
= \mathbb{E}_{z}\!\Big[ -\log\sigma\!\big(
\mathbf{t}_q^\top(\mathbf{t}_+ - \mathbf{t}_-) + z \big) \Big].
\label{eq:l-t2v-noise}
\end{equation}
\textbf{Step 3: Proving the lower bound}
We will use Jensen's inequality, \ie, given a convex function $f$ and random variable $X$, Jensen's inequality states:
\begin{equation}
    \label{eq:jensen}
    f\left(\mathbb{E}[X] \right) \leq E\left[ f(X) \right].
\end{equation}
Notice that $f := - \log \sigma $ is a convex function. Thus, applying \cref{eq:jensen} to \cref{eq:l-t2v-noise}, we get
\begin{align*}
    \ell^{t2v} &= \mathbb{E}_{z}\!\Big[ -\log\sigma\!\big(
\mathbf{t}_q^\top(\mathbf{t}_+ - \mathbf{t}_-) + z \big) \Big] \\
    &\geq -\log \sigma \Big(  \mathbb{E}_{z}\big[ t_{q}^{T}(t_{+} - t_{-}) + z \big]  \Big) \\
    &= -\log \sigma \Big(  t_{q}^{T}(t_{+} - t_{-}) + \cancelto{0}{\mathbb{E}_{z}[z]}  \ \Big) \\
    &= -\log \sigma \big(t_{q}^{T}(t_{+} - t_{-}) \big) \\
    &= \ell^{text-only}.
\end{align*}
This completes the proof.
\end{proof}

\begin{proposition}[Upper bound]
\label{prop:noise-bounds}
Under the MGA, with unit-norm query embeddings, the text-to-video loss of
any triplet, after averaging over the alignment noise, satisfies
\begin{equation}
\ell^{t2v} \;\le\; e^{\nu^2}\, \ell^{\text{text-only}}.
\label{eq:multiplicative}
\end{equation}
This holds pointwise per triplet with triplet-independent
constants, hence also between the population losses:
$\mathcal{L}_{t2v} \le e^{\nu^2}\, \mathcal{L}_{\text{text-only}}$.
\end{proposition}

The proof shares the Steps 1 and 2 from that of \cref{prop:lower-bound}. We detail the remaining steps below.

\begin{proof}[Proof of Proposition~\ref{prop:noise-bounds}]
\\[2mm]
\textbf{Step 3: Rewrite the sigmoid loss in exponential form.} Since
$\sigma(x) = (1 + e^{-x})^{-1}$, we have
$-\log\sigma(x) = \log(1 + e^{-x})$. Applying this to
\eqref{eq:l-t2v-noise} and splitting the exponent via
\eqref{eq:margin-split},
\begin{equation}
\ell^{t2v}
= \mathbb{E}_z\!\left[ \log\!\Big\{ 1 +
\exp\!\big( -\mathbf{t}_q^\top(\mathbf{t}_+ - \mathbf{t}_-)\big)
\cdot \exp(-z) \Big\} \right],
\label{eq:exp-form}
\end{equation}
so the randomness enters only through the multiplicative factor
$\exp(-z)$.
\\[2mm]
\textbf{Step 4: Average out the noise via Jensen's inequality and the
Gaussian moment generating function.} For fixed text margin, the
expression inside the expectation in \eqref{eq:exp-form} is $\log$ of an
\emph{affine} function of the random quantity $\exp(-z)$. Since
$\log(\cdot)$ is concave, Jensen's inequality gives
$\mathbb{E}[\log Y] \le \log \mathbb{E}[Y]$, hence
\begin{equation}
\ell^{t2v}
\;\le\;
\log\!\Big\{ 1 +
\exp\!\big(-\mathbf{t}_q^\top(\mathbf{t}_+ - \mathbf{t}_-)\big)\cdot
\mathbb{E}_z\big[\exp(-z)\big] \Big\},
\label{eq:jensen-upper}
\end{equation}
using linearity of expectation. By the Gaussian moment generating
function, if $x \sim \mathcal{N}(0, s^2)$ then
$\mathbb{E}_x[\exp(-x)] = e^{s^2/2}$; with \eqref{eq:z-law} this gives
$\mathbb{E}_z[\exp(-z)] = e^{2\nu^2/2} = e^{\nu^2}$. Substituting into
\eqref{eq:jensen-upper} yields the common intermediate bound
\begin{equation}
\ell^{t2v} \;\le\; \log\!\Big\{ 1 +
\exp\!\big(-\mathbf{t}_q^\top(\mathbf{t}_+ - \mathbf{t}_-)\big)\,
e^{\nu^2} \Big\}.
\label{eq:intermediate}
\end{equation}
\\[2mm]
\textbf{Step 5: Multiplicative bound.} We use the following elementary
inequality: for every $c \ge 0$ and every $A \ge 1$,
\begin{equation}
\log(1 + Ac) \;\le\; A \log(1 + c).
\label{eq:elementary}
\end{equation}
To see this, fix $c \ge 0$ and view both sides as functions of $A$ on
$[1,\infty)$. At $A = 1$ they are equal. The left side has derivative (w.r.t. $A$)
$c/(1 + Ac)$ in $A$, while the right side has constant derivative
$\log(1+c)$. Since $A \ge 1$,
\[
\frac{c}{1 + Ac} \;\le\; \frac{c}{1 + c} \;\le\; \log(1 + c),
\]
where the last step is the standard bound
$\log(1+c) = \int_0^c \frac{dx}{1+x} \ge \int_0^c \frac{dx}{1+c}
= \frac{c}{1+c}$. Thus the right side of \eqref{eq:elementary} grows at
least as fast as the left everywhere on $[1,\infty)$, and they start
equal, proving \eqref{eq:elementary}.

Applying \eqref{eq:elementary} to \eqref{eq:intermediate} with
$c = \exp\!\big(-\mathbf{t}_q^\top(\mathbf{t}_+ - \mathbf{t}_-)\big)
\ge 0$ and $A = e^{\nu^2} \ge 1$,
\[
\ell^{t2v}
\;\le\;
e^{\nu^2} \log\!\Big\{ 1 +
\exp\!\big(-\mathbf{t}_q^\top(\mathbf{t}_+ - \mathbf{t}_-)\big) \Big\}
= e^{\nu^2}\, \ell^{\text{text-only}},
\]
which proves \eqref{eq:multiplicative}.
\end{proof}

\begin{remark}[Sandwich]
Combined with the lower bound (\cref{prop:lower-bound}), the two directions give
\[
\ell^{\text{text-only}}
\;\le\; \ell^{t2v} \;\le\;
e^{\nu^2}\, \ell^{\text{text-only}}.
\]
In particular
$1 \le \mathcal{L}_{t2v} / \mathcal{L}_{\text{text-only}} \le e^{\nu^2}$,
so driving the text-only loss to zero drives the text-to-video loss to
zero at the same rate up to the constant $e^{\nu^2}$.
\end{remark}

\begin{remark}[Without normalization]
If query embeddings are not unit-norm, every step goes through with
$\nu^2$ replaced by $\nu^2 \|\mathbf{t}_q\|^2$ per triplet; a uniform
population bound then holds with
$\nu^2 \sup_q \|\mathbf{t}_q\|^2$ in place of $\nu^2$.
\end{remark}

\begin{remark}[Softmax temperature]
    If the contrastive loss involves temperature $\tau$, then the upper bound is modified to be $e^{\nu^2 / \tau^2}$. 
\end{remark}
\begin{remark}[Empirical bound]
    We experimented with the base model Tarsier-2's embeddings for video-text pairs in MSRVTT. Assuming $\mathbf{\epsilon} \sim \mathcal{N}(\mathbf{0}, \nu^{2}\mathbf{I})$, we empirically observe $\nu \approx 0.03$ and the softmax temperature is $\tau = 0.05$ which leads to the bounds: $1 \leq \mathcal{L}_{\text{t2v}}/\mathcal{L}_{\text{text}} \leq \exp(\nu^{2}/\tau^{2}) = 1.17$. While the theoretical bound is valid, empirically, we find that $\mathbf{\epsilon}$ is not necessarily a diagonal Gaussian which leads to a looser bound of $1.35$.
\end{remark}

\begin{remark}[Generalization to approx. Gaussian noise]
In case $\mathbf{\epsilon}$ is not a zero cross-covariance Gaussian, the bound can be generalized as long as one can compute the term $\mathbb{E}_{z}[\exp(-z)]$ since that is the only place we used the Gaussian assumption.
\end{remark}

\begin{figure}[h]
\captionsetup{skip=-1mm}
    \centering
    \includegraphics[width=\linewidth]{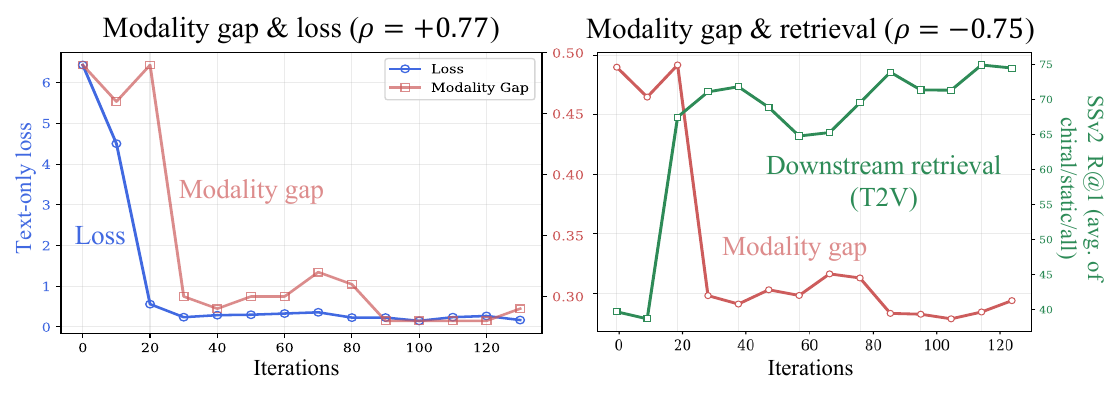}
    \caption{Empirical measurement of modality gap over training. (left) The modality gap decreases along with the training loss. (right) The downstream retrieval performance improves as the modality gap decreases.}
    \label{fig:modgap-all}
\end{figure}

\asubsection{Augmenting with experimental evidence.} 
\label{aubsec:modgap-experiments}
We also augment this theoretical analysis empirically with several experiments. (1) We compute the \textbf{modality gap} ($\lVert\Delta_{\text{gap}}\rVert$ 
from~\cite{zhang2023diagnosing}) over training iterations in \cref{fig:modgap-all}. The loss decreases in tandem with $\lVert\Delta_{\text{gap}}\rVert$ with a correlation of $\rho=0.77$; the downstream retrieval (SSv2) increases as $\lVert\Delta_{\text{gap}}\rVert$ decreases ($\rho{=-}0.75$). (2) We also measure the \textbf{uniformity pressure loss} ($\mathcal{L}_{\text{uniform}}$ defined 
in~\cite{wang2020understanding}) 
in each modality. In both modalities, TARA reduces uniformity loss compared to the base model (text: $-1.07 \rightarrow -3.3$, video: ${-1.2} \rightarrow {-3.4}$), i.e., the embeddings are more spread apart in TARA.

\begin{figure}[h]
    \centering
    \includegraphics[width=\linewidth]{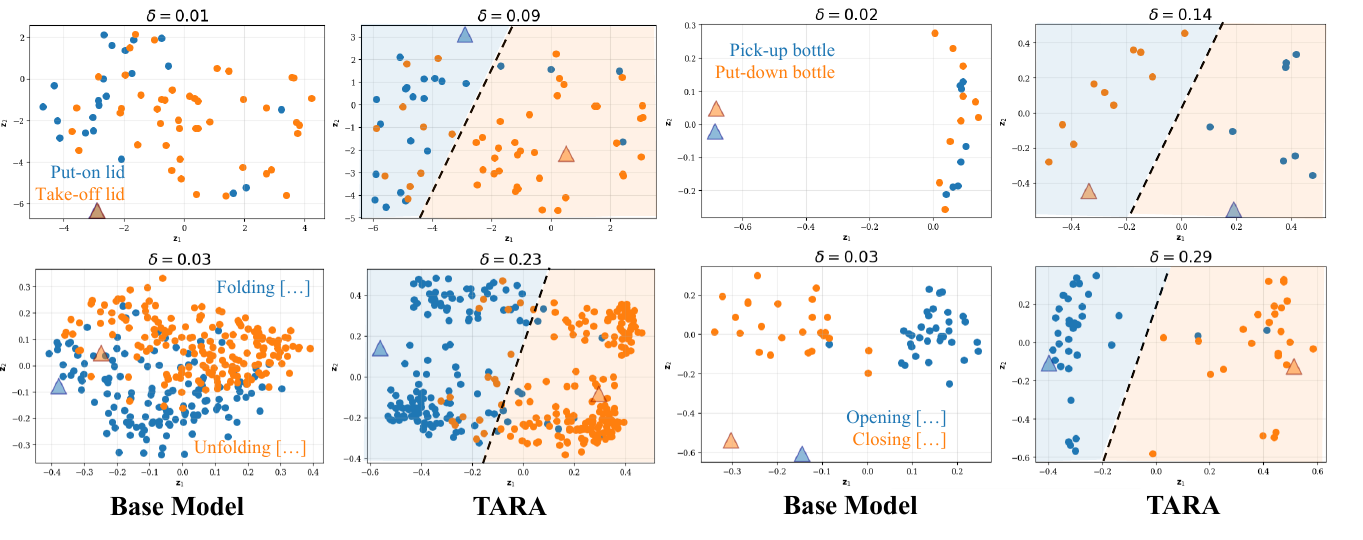}
    \caption{\textbf{Visualizing embeddings for videos of chiral actions.} Following up on Fig.~6 in the main paper, we show more chiral action pairs where (i) video embeddings are better separable, and (ii) text embeddings are better aligned with their video clusters post \TARA fine-tuning.
    Text embeddings are shown as $\triangle$.  $\delta$ measures the avg.\
similarity of text embedding with its matched videos. A higher $\delta$ indicates better alignment.}
    \label{fig:supp-tsne-chiral}
\end{figure}

\asection{Qualitative Results}
\label{sec:qual}
\noindent\textbf{Visualizing embeddings.} Following up on Fig.~6 in the main paper, we show more chiral action pairs where (i) video embeddings are better separable, (ii) text embeddings are better aligned with their video clusters post \TARA fine-tuning in \cref{fig:supp-tsne-chiral}.
\vspace{2mm}

\begin{figure}[h]
    \centering
    \includegraphics[width=0.8\linewidth]{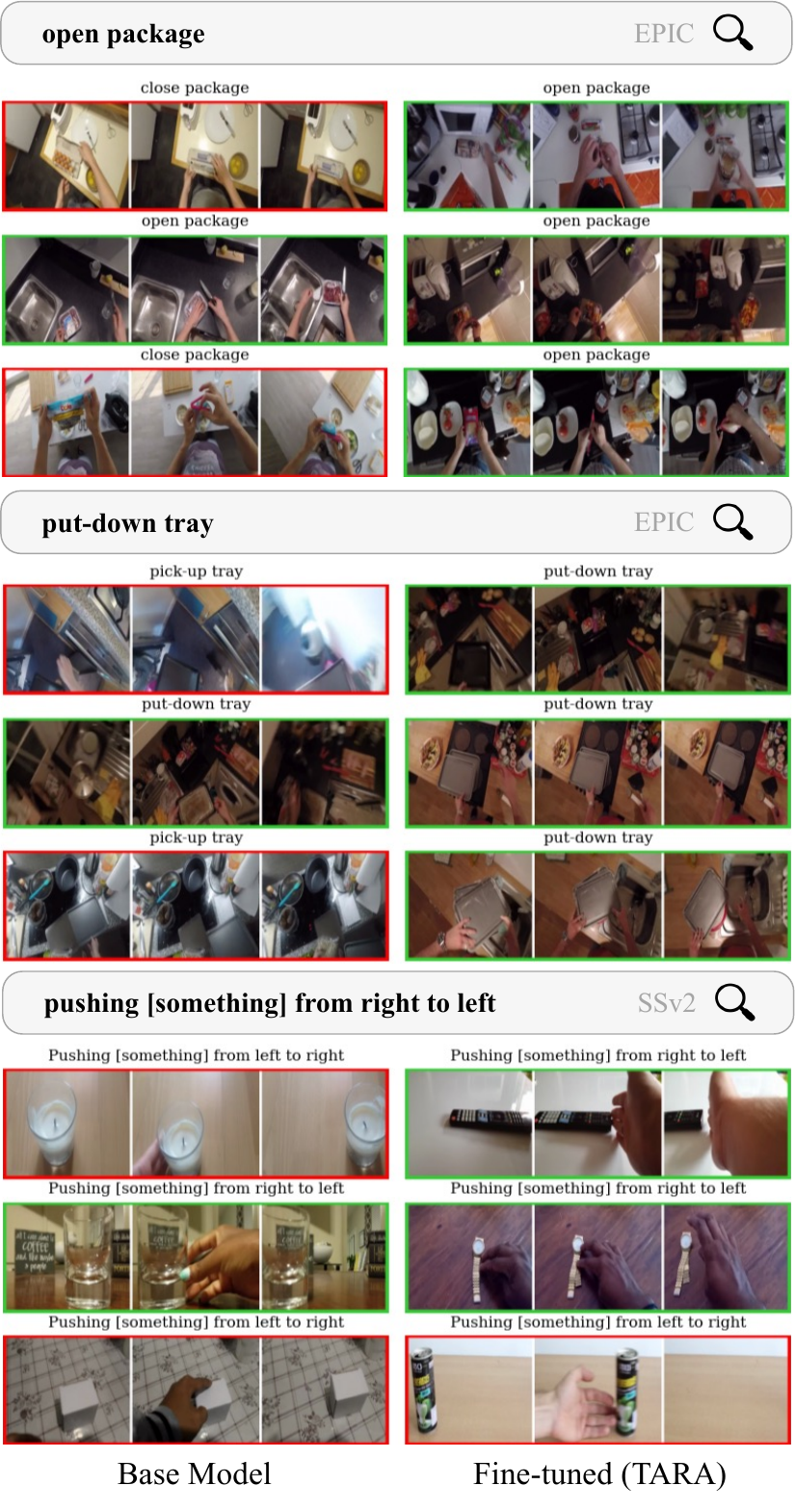}
    \caption{\textbf{Retrieval results for chiral actions.} \TARA yields qualitatively better results for queries involving  chiral actions (ones with temporally opposite actions). Top 1 results is marked by red (incorrect) or green (correct).}
    \label{fig:qual-chiral}
\end{figure}

\begin{figure}[h]
    \centering
    \includegraphics[width=\linewidth]{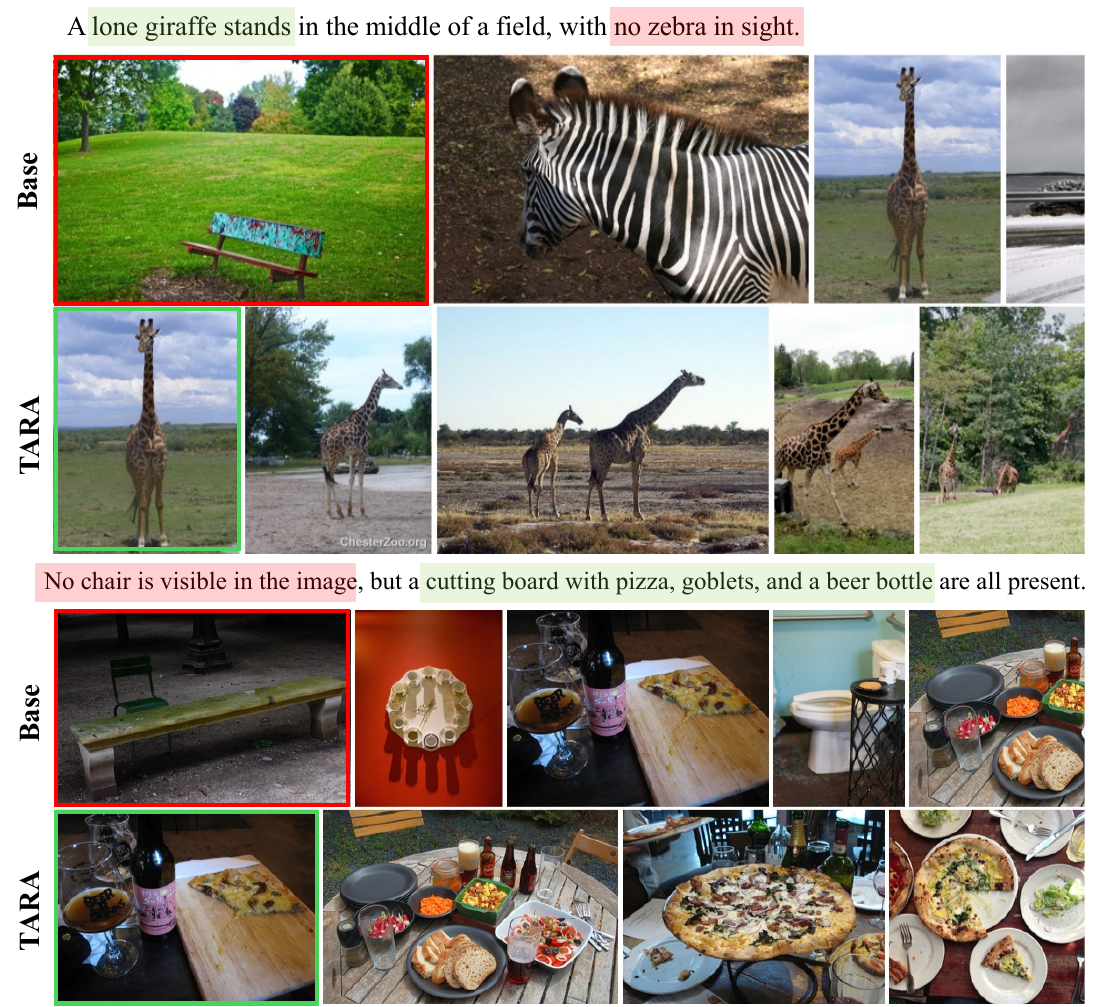}
    \caption{\textbf{Retrieval results with negation in queries.} Base model is distracted by negation, while \TARA fine-tuned model retrieves accurately. Top 1 results is marked by red (incorrect) or green (correct).}
    \label{fig:qual-neg}
\end{figure}

\begin{figure}[h]
    \centering
    \includegraphics[width=\linewidth]{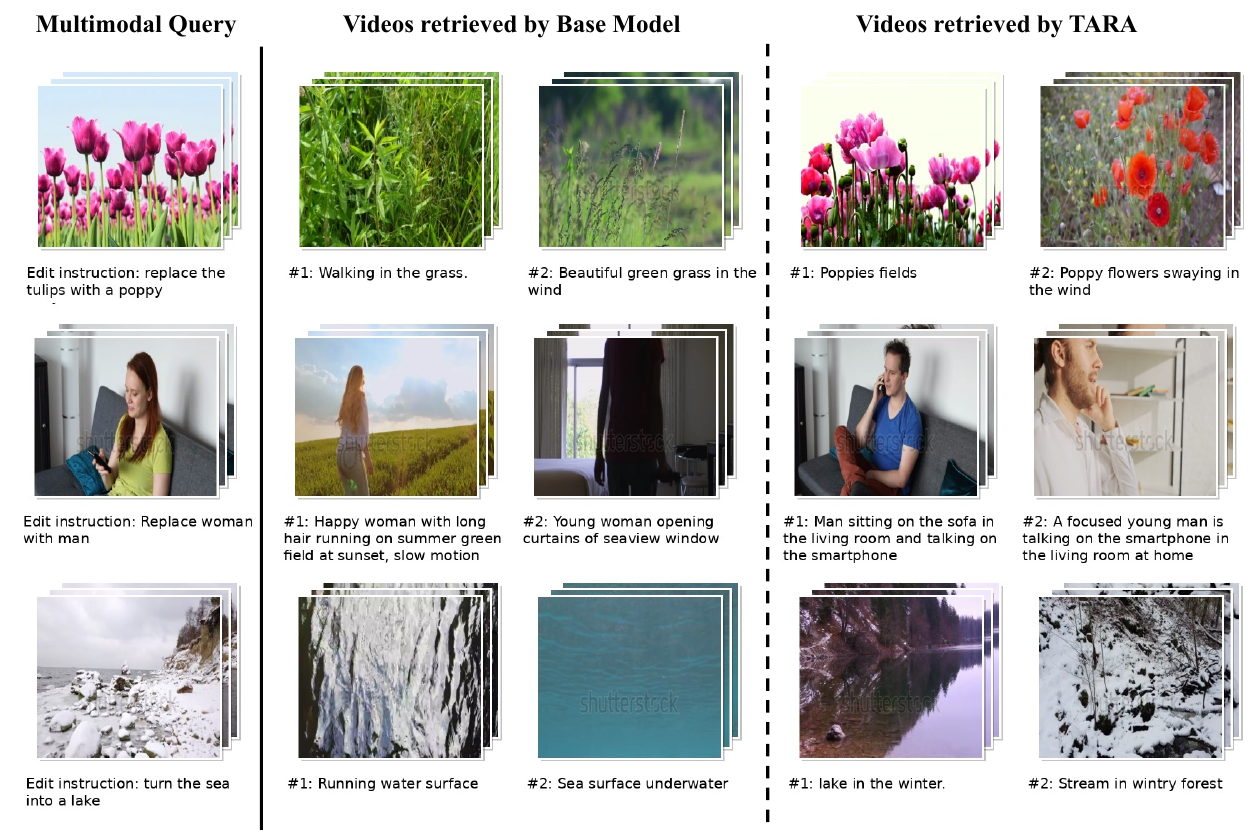}
    \caption{\textbf{Retrieval results for Composed queries.} (Left) Input query composed of a video and edit instruction. Top-2 retrieved videos from base model (middle) and \TARA (right). Captions of candidates are not used.}
    \label{fig:qual-covr}
\end{figure}

\noindent\textbf{Retrieval results.} Following up on Fig.~3 in the main paper, we present more qualitative retrieval results for each of the three nuances: (i) temporal (\cref{fig:qual-chiral}), (ii) negation (\cref{fig:qual-neg}) and (iii) multimodal (\cref{fig:qual-covr}).
\vspace{2mm}


\noindent\textbf{Visualizing logits.} Following up on Fig.~5 in the main paper, we visualize two more video-caption pairs for which we visualize the predicted logits before and after \TARA fine-tuning.

\begin{figure}[h]
    \centering
    \includegraphics[width=\linewidth]{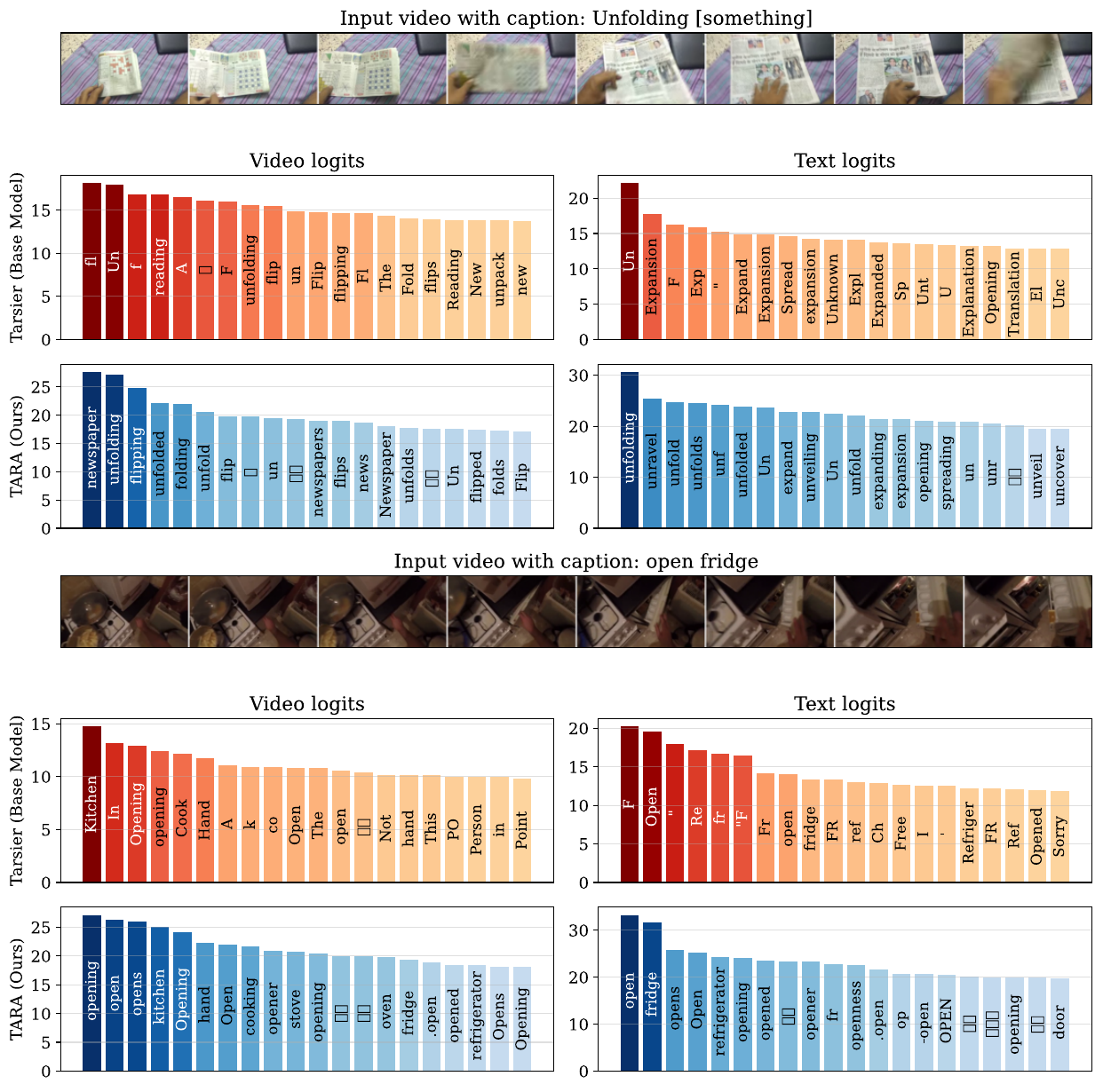}
    \caption{\textbf{Visualizing logits from generated embeddings.} Following up on Fig.~5 in the main paper, we visualize two more video-caption pairs for which we visualize the predicted logits before and after \TARA fine-tuning.}
    \label{fig:logits}
\end{figure}

\end{document}